\documentclass{article} 
\usepackage[toc,page,header]{appendix}
\usepackage{iclr2025_conference,times}
\usepackage{minitoc}
\usepackage{graphicx}
\usepackage{wrapfig}
\usepackage{amsthm}
\usepackage{float}

\usepackage{amsmath,amsfonts,bm}

\def\eqref#1{equation~\ref{#1}}

\def\1{\bm{1}}

\DeclareMathAlphabet{\mathsfit}{\encodingdefault}{\sfdefault}{m}{sl}
\SetMathAlphabet{\mathsfit}{bold}{\encodingdefault}{\sfdefault}{bx}{n}

\usepackage{listings}
\usepackage{xcolor}
\usepackage{hyperref}
\usepackage{url}

\title{VSF: Simple, Efficient, and Effective Negative Guidance in Few-Step Image Generation Models By \underline{V}alue \underline{S}ign \underline{F}lip}

\author{Wenqi Guo \thanks{Also affilated to Department of MEOW, Weathon Software}  \& Shan Du \thanks{Corresponding author} \\
Department of CMPS\\
University of British Columbia\\
Kelowna, BC V1V 1V8, Canada \\
\texttt{wg25r@student.ubc.ca,shan.du@ubc.ca} \\
}

\iclrfinalcopy 
\begin{document}

\maketitle
\doparttoc 
\faketableofcontents 

\begin{abstract}
We introduce Value Sign Flip (VSF), a simple and efficient method for incorporating negative prompt guidance in few-step (1-8 steps) diffusion and flow-matching image and video generation models. Unlike existing approaches such as classifier-free guidance (CFG), NASA, and NAG, VSF dynamically suppresses undesired content by flipping the sign of attention values from negative prompts. Our method requires only a small computational overhead and integrates effectively with MMDiT-style architectures such as Stable Diffusion 3.5 Turbo and Flux Schnell, as well as cross-attention-based models like Wan. We validate VSF on a challenging dataset, NegGenBench, with complex prompt pairs. Experimental results on our proposed dataset show that VSF significantly improves negative prompt adherence (reaching 0.420 negative score for quality settings and 0.545 for strong settings) compared to prior methods in few-step models (scored 0.320-0.380 negative score) and even CFG in non-few-step models (scored 0.300 negative score), while maintaining competitive image quality and positive prompt adherence. Our method is also a suppressed generate-then-edit pipeline, while also having a much faster runtime. Code, ComfyUI node, and dataset are available in \url{https://github.com/weathon/VSF/tree/main}.

\end{abstract}
\begin{figure*}[h]
    \centering
    \includegraphics[width=0.8\linewidth]{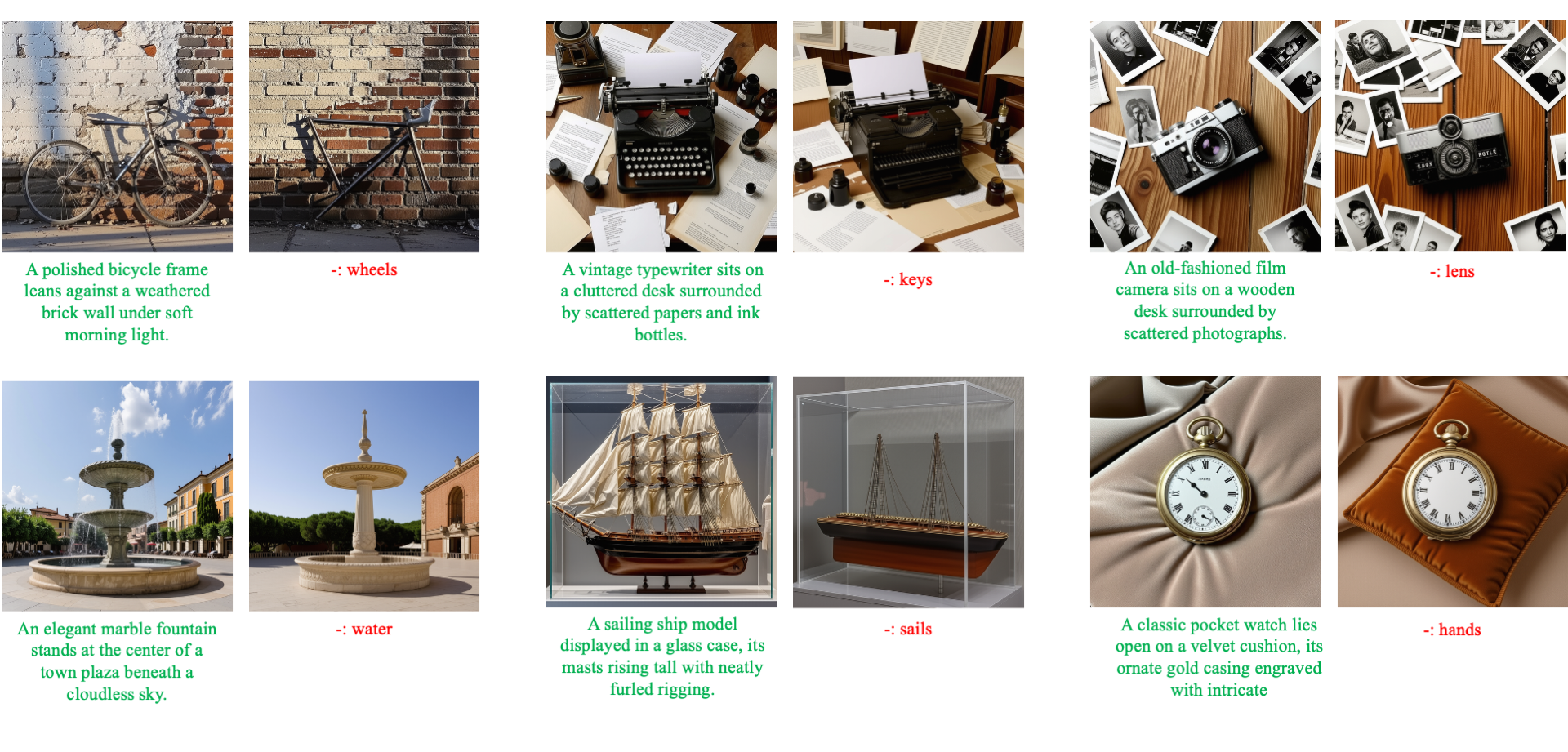}
    \caption{Original image without negative guidance and image generated using our VSF negative guidance on Stable Diffusion 3.5 Large Turbo. The green prompt is the positive prompt, and the red one is the negative prompt. These examples have significant challenges as they are removing essential parts of an object. The ``hands'' in the last image mean clock hands.} 
    \label{fig:main}
\end{figure*}
\section{Introduction}
Diffusion models (including flow matching models) have demonstrated their ability to produce diverse and high-quality images \citep{noauthor_black-forest-labs/flux.1-schnell_2025, noauthor_stable_2022, sd3} and videos \citep{wan_wan:_2025, yin_slow_2025}. However, a longstanding issue remains: the challenge of effectively applying negative guidance in image and video generation. Addressing this problem is crucial for improving content control, moderation \citep{schramowski_safe_2023}, quality assurance, and reducing biases when generating general concepts \citep{chen_normalized_2025}. However, vision language models (VLMs) have difficulties interpreting negations \citep{park_know_2025, alhamoud_vision-language_2025, singh_learning_2025, singh_learn_2024}, rendering prompts containing negations ineffectively or made the negative prompt appears even more (e.g., a prompt like “a scientist who is not wearing glasses” will often generate a scientist with glasses—sometimes even more frequently than a simple prompt like “a scientist”). Classifier-free guidance (CFG) \citep{ho_classifier-free_2022} can be used to address this issue when substituting unconditional generation with negative guidance.

\begin{wrapfigure}{r}{0.3\textwidth}
    \centering
    \includegraphics[width=0.8\linewidth]{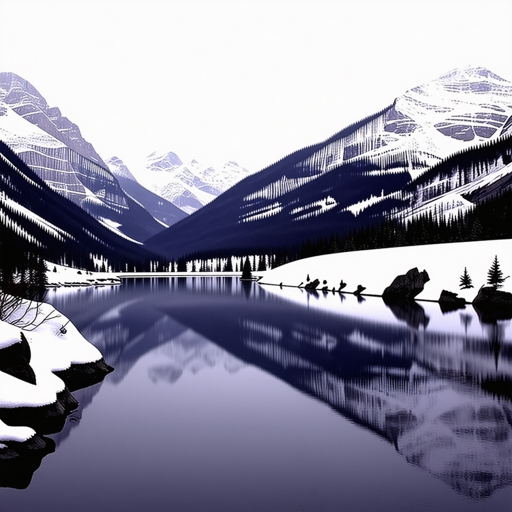}
    \caption{An example of forcefully applying CFG to a step-distilled model is shown using a guidance scale of 2.8 and only 4 steps on SD-3.5-Large Turbo. The positive prompt describes a Canadian winter landscape, while the negative prompt includes the word “lake.” The resulting image does not cancel the lake and exhibits severe over-saturation artifacts (trees in the background).} 
    \label{fig:bad_cfg}
\end{wrapfigure}

However, to enhance efficiency in image and video generation, numerous models have been distilled to support inference in just a few steps (1-8 steps), such as Flux Schnell \citep{noauthor_black-forest-labs/flux.1-schnell_2025}, Stable Diffusion 3.5 Large Turbo \citep{sd3}, SDXL Lighting \citep{lin_sdxl-lightning:_2024}, SNOOPI \citep{nguyen_snoopi:_2024}, and CausVid \citep{noauthor_kijai/wanvideo_comfy_nodate, yin_slow_2025}. However, CFG is incompatible with these models. These models are usually distilled and run in CFG-disabled mode, which means only the positive guidance is used, and there is no extrapolation. When CFG is applied forcefully, the resulting image often becomes oversaturated, particularly when the CFG scale is set high enough to suppress unwanted concepts. Moreover, if the number of diffusion steps is too low, the output may reflect features from both the positive and negative prompts \citep{nguyen_snoopi:_2024}, rather than excluding the negative prompt entirely. This occurs due to a divergence between the positive and negative guidance signals \citep{chen_normalized_2025}. An example is shown in Figure~\ref{fig:bad_cfg}. Additionally, even if CFG works, it requires two forward passes, one for positive guidance and one for negative guidance, which doubles the run time.

To address this, two methods, Negative Steer Away Attention (NASA) \citep{nguyen_snoopi:_2024} and Normalized Attention Guidance (NAG) \citep{chen_normalized_2025}, have been introduced, employing negative guidance within attention final output space rather than the output space. NASA is currently limited to cross-attention models (though it can be re-implemented into other models), while NAG primarily targets quality control rather than negative prompt avoidance. Both methods calculate positive and negative attentions separately and subtract them using a prefixed scale (same as CFG), resulting in a fixed guidance strength throughout the generation, across different areas of the image, and at different layers of the model. This approach lacks adaptability to various time steps, layers, or image regions, limiting effectiveness in negative prompt adherence compared to a more adaptive method \citep{schramowski_safe_2023, koulischer_dynamic_2025, ban_understanding_2024}.

In this study, we introduce Value Sign Flip (VSF), a method that dynamically adjusts the guidance strength by flipping the sign of negative prompt values \textit{within} the attention calculation (i.e., not the attention output). This enables the model to steer away from negative concepts adaptively based on their current presence strength, similar to the approach of \citet{koulischer_dynamic_2025}. VSF has a small computational overhead and, when combined with few-step models, facilitates extremely fast image or video generation (<3 seconds). Our contributions in this work are: (1) we proposed a new method for better negative guidance; (2) we constructed a dataset, NegGenBench, consisting of challenging positive-negative prompt pairs; (3) we collected images generated using these three methods (VSF, NAG, NASA) and labeled their negative and quality score. We further fine-tuned a VLM on it for future work to better evaluate negative prompt following.

\section{Related Work}
\subsection{Classifier Free Guidance} 
Vision language models struggle to understand negation \citep{yuksekgonul_when_2023, singh_learn_2024, alhamoud_vision-language_2025, park_know_2025} (We discussed more about this in the Appendix). Original classifier-free guidance (CFG) \citep{ho_classifier-free_2022} generates a conditioned noise prediction and an unconditioned noise prediction. In flow matching \citep{lipman_flow_2023}, the predicted targets are the velocity ($u_t$). Thus, the original flow matching CFG prediction can be written as 
\begin{equation}
u_t = f(\emptyset, x_{t+1}, t) + \lambda (f(p^+, x_{t+1}, t) - f(\emptyset, x_{t+1}, t)),
\end{equation}
where $p^+$ is the positive prompt, $x_t$ is the latent at time $t$ (where higher $t$ means more torward the noise distribution), $f(\cdot)$ is the trained model, and $\lambda$ is the guidance scale.  Later, the community finds out that by replacing the unconditional generation with a negative prompt (e.g., description of an unwanted image), the model will avoid the prompt due to the negative sign. This is the common implementation of a negative prompt. This turns the above equation into 
\begin{equation}
    u_t = f(p^-, x_{t+1}, t) + \lambda (f(p^+, x_{t+1}, t) - f(p^-, x_{t+1}, t)) ,
\end{equation} 
where $p^-$ is the negative prompt.

\subsection{Recent Works on Dynamic Negative Guidance}
The studies on dynamic negative guidance are very limited (only \citep{ban_understanding_2024, koulischer_dynamic_2025, schramowski_safe_2023}). Ban \textit{et al.} \citep{ban_understanding_2024} found that the negative prompts affect the model by delayed effects and neutralization. After the model has generated unwanted contents, the negative guided output ($u_{p^-}$) will neutralize the content. They also observed the reverse activation effect, where the negative prompt introduced early in the diffusion processes could actually induce the unwanted concepts. To address this, they proposed applying the negative guidance later in the diffusion process and found it effective. 

 \citet{schramowski_safe_2023} used a very similar idea as CFG to avoid unwanted (NSFW) content. They generate an unsafe vector and purposely avoid it by subtracting it from the predicted noise. They also added a pixel-level guidance scale that depends on the pixel-wise distance between the positive predicted noise and the unwanted noise, making it adaptive to different regions in the image.

 \citet{koulischer_dynamic_2025} used similar ideas of both and proposed a temporal dynamic guidance scale method. They calculate a probability that the generated concept contains negative content and adjust the guidance scale accordingly. However, their adaptive scale only changes throughout the steps and does not adapt to different regions in the image. 
 

\subsection{Few-Step Image Generation Models}
Traditional diffusion or flow-matching image generation models typically require many inference steps. However, with improved sampler, this can be reduced to around 20 steps. Recent approaches go further by using step distillation to reduce the number of steps to fewer than 8, or even a single step, as demonstrated in Flux Schnell~\citep{noauthor_black-forest-labs/flux.1-schnell_2025}, SDXL Lightning~\citep{lin_sdxl-lightning:_2024}, CausVid \citep{noauthor_kijai/wanvideo_comfy_nodate, yin_slow_2025}, SNOOPI \citep{nguyen_snoopi:_2024}, and Stable Diffusion 3.5 Turbo~\citep{sd3}. Since these models are distilled, they generally do not use classifier-free guidance (CFG) during inference; when CFG is forcibly applied, the results are significantly degraded to the point that it is completely unusable~\citep{nguyen_snoopi:_2024}, see Figure~\ref{fig:bad_cfg} for an example. 

\subsection{Recent Works on Negative Guidance in Few-Step Models}
Recently, two approaches have specifically targeted negative guidance techniques for few-shot models: Negative-Away Steer Attention (NASA) \citep{nguyen_snoopi:_2024} and Normalized Attention Guidance (NAG) \citep{chen_normalized_2025}. Although they both focused on avoiding unwanted content and improving quality (using a negative prompt that describes bad quality), NASA mainly focused on avoiding unwanted content, while NAG focused on improving quality. 

The authors of the NASA study found that neither standard CFG nor CFG applied directly to text embeddings yields desirable results in few-step scenarios, particularly in single-step settings. Specifically, the regular CFG independently computes positive and negative guidance signals, preventing the negative guidance from effectively neutralizing unwanted concepts. As a result, the produced images merely appear as a mixture of both positive and negative prompts unnaturally (an average image of the positive prompt generated image and the negative prompt independently generated image) rather than excluding negative prompt elements. Furthermore, the authors noted that applying CFG to text embeddings produces minimal benefits. For detailed examples and further illustration, readers could refer to the original paper introducing NASA \citep{nguyen_snoopi:_2024}.

NASA applies the guidance in intermediate states (attention outputs) instead of the final predicted noise or velocity. Specifically, they calculate a positive attention output $Z^+$ and a negative attention output $Z^-$, and the final attention $Z^{NASA}$ is obtained by subtracting the two with a factor $\alpha$, as shown in Equation~\ref{eq:NASA}. The alpha value is usually between 0 and 1. 
\begin{equation}
    \label{eq:NASA}
\end{equation}

\begin{wrapfigure}{r}{0.5\textwidth}
    \centering
    \includegraphics[width=0.9\linewidth]{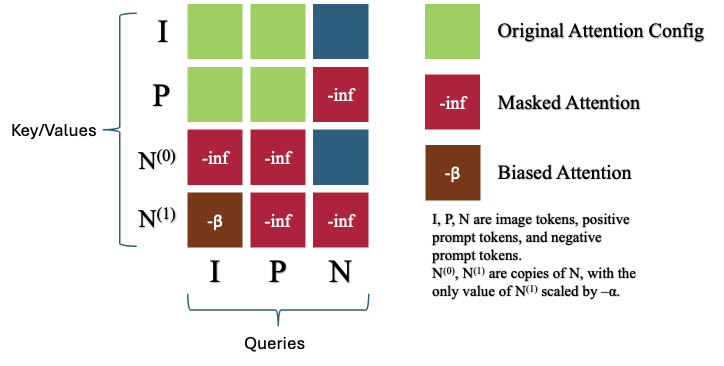}
    \caption{The attention mechanism of our method. We pass in image tokens ($I$), positive prompt tokens ($P$), and negative prompt tokens ($N$) into attention. For key and values, $N$ is duplicated, with values of one copy ($N^{(1)}$) scaled by $-\alpha$. Some areas are masked to avoid interference. An bias $-\beta$ is added to $I\rightarrow N^{(1)}$ attention.}
    \label{fig:method}
    \vspace{0.5cm}
\end{wrapfigure}

Normalized Attention Guidance (NAG) used a similar approach. But instead of subtracting the negative attention map from the positive, it uses a similar extrapolation approach as CFG, as shown in Equation~\ref{eq:nag1}. The starting point $Z^+$ could also be replaced with $Z^-$; they are equivalent if $\phi$ is increased by 1.
\begin{equation}
    \label{eq:nag1}
    \widetilde{Z}^{NAG} = Z^+ + \phi (Z^+ - Z^-)
\end{equation}
However, to maintain the stability of the attention output space, they also applied normalization to $\widetilde{Z}^{NAG}$ to limit its norm relative to $Z^+$ with scale $\tau$ per token, resulting in $\hat{Z}$. Then it used a blending factor $\alpha$ to blend it with the positive attention result, as shown in Equation~\ref{eq:nag3}.
\begin{equation}
    \label{eq:nag3}
    Z^{NAG} = \alpha\hat{Z} + (1-\alpha) Z^+
\end{equation}
The normalization and blending ensure the attention output of the NAG does not drift away from what the model usually sees during training, improving the quality of generated images. However, if the constraint is set to be too tight (i.e., high $\alpha$ and low $\tau$), it might also limit the model's ability to follow negative prompts. 

\subsection{Other Related Works}
There is previous work that controls attention to mainlutplate images. Attend-and-Excite \citep{chefer_attend-and-excite_2023} forces the attention on all key tokens to avoid the generation missing some terms in the prompt. Self-Guidance \citep{epstein_diffusion_2023} mantluplate attention to changing elements' properties in the image. BoxDiff \citep{xie_boxdiff_2023} controls the attention map such that the object appears on the desired location in the image. However, our work is different in that we used it in a way to cancel unwanted elements by simply flipping the sign of values in the negative prompt.

Our work is also related to debiasing in image generation, since it can be used as a way to move away from learned associations (which could be about demography bias or object pairings). Previous work mainly targeted on the input, such as using reference images as input \citep{zhang_iti-gen_2023} or using a learned prompt. FairQueue \citep{teo_fairqueue_2024}. Our method is different such that we control the value in the negative prompt. Additionally, our method can work against very strong associations like a bike without wheels.

\section{Proposed Methods}
Our proposed method is built on top of NASA \citep{nguyen_snoopi:_2024},  \citet{koulischer_dynamic_2025}, and \citet{schramowski_safe_2023}. NASA has a fixed guidance for every attention calculation, \citet{koulischer_dynamic_2025} does not have token-level modulation, and \citet{schramowski_safe_2023}'s approach of taking the differences between the positive and negative noise predictions is too simple for complex situations, as mentioned in \citet{koulischer_dynamic_2025}. In NAG's \citep{chen_normalized_2025} future work section, they also mentioned the possibility of token-level modulation but they did not propose a specific solution. 

\subsection{Value Sign Flip Adaptive Attention}

We propose to expand \citet{koulischer_dynamic_2025} idea to token-level modulation in few-step models. Let $W$ be a per-token weight at each attention calculation for how strongly the token is associated with a positive concept compared to the negative one. We can modify the NASA attention to Eq.~\ref{eq:ours1}. $W$ is obtained by a function with a positive prompt, a negative prompt, and an image as input. 

\begin{equation}
    W=g(p^+, p^-, I),
\end{equation}
then we can rewrite the equation in NASA as
\begin{equation}
    \label{eq:ours1}
   Z^{W} = WZ^+ - \alpha (1-W)Z^- 
\end{equation}

An intuitive method to calculate $W$ involves using the model’s attention map: when the image attends more to the negative prompt compared to the positive one, it should be steered away strongly accordingly. Thus, we can calculate the attention map between the image and the positive tokens $A^+$ and the image and the negative tokens $A^-$ before softmax calculation, then calculate their ratios to their sum. $Q$ is the image query tokens and $K^+$ and $K^-$ are the positive and negative prompt keys.

\begin{equation}
    A^+=\exp(\frac{Q(K^+)^T}{\sqrt{d}}), A^-=\exp(\frac{Q(K^-)^T}{\sqrt{d}}),
W = \frac{\sum A^+ }{\sum(A^++A^-)}
\end{equation} 
This approach involves complex attention calculation and two attention passes, but it can be implemented by a simpler approach. We can concatenate the values and keys of the positive and negative prompts, then flip the sign of the negative prompt values. This enables that when the image attends to the negative prompt, the flipped value of the negative prompt can cancel the unwanted content. The equation of our method in cross attention models, written in the matrix calculation, is shown in Equation~\ref{eq:ours}, where $\oplus$ means matrix concatenation on the sequence length dimension, $\sigma$ is the softmax function on the sequence length dimension, and $V^+$ and $V^-$ are the positive and negative prompt values.

\begin{equation}
    \label{eq:ours}
    Z^{VSF} = \sigma (\frac{Q(K^+\oplus K^-)^T}{\sqrt{d}}) (V^+\oplus-\alpha V^-)
\end{equation} 
This is similar to noise-canceling headphones, where a ``flipped'' wave is played to cancel the noise. Note that the key of the negative prompt is not flipped to keep the original meaning of the unwanted concept to match image patches. Mathematically, this is equivalent to $Z^W$. Proof is in the Appendix.

This approach gives a dynamic weight for the positive and negative prompts, and it varies for different layers, steps, and tokens. 

\subsection{Attention Masking and Duplication of Negative Embedding}
The above method works well for cross-attention-based methods, where attention only exists between image-to-image in self-attention layers and image-to-text in cross-attention layers. However, it requires modification, including masking and duplication, to work in MMDiT-style models such as SD3.5~\citep{sd3}, where all image and text tokens are concatenated into a single sequence before attention.

In the standard MMDiT-style setup without our method (e.g., using CFG, NASA, or NAG), the sequence inputs for the attention module are:
$
[\mathbf{I},\ \mathbf{P}]\ \text{and}\ [\mathbf{I},\ \mathbf{N}].
$
If we concatenate all tokens into a single sequence without any modification, we will get: $
[\mathbf{I},\ \mathbf{P},\ \mathbf{N}],
$ where $\mathbf{I}$ represents image tokens, $\mathbf{P}$ represents positive prompt tokens, and $\mathbf{N}$ is the negative prompt. During attention, queries, keys, and values are all projected from this combined sequence.

If we apply a sign flip to the negative prompt values by scaling $V_N=V\mathbf{N}$ with $-\alpha$ (where $V$ is the value projection), this flipped content affects all attention paths involving $V_{\mathbf{N}}$. That includes not only the intended interaction between image and negative prompt $(\mathbf{I} \rightarrow \mathbf{N})$\footnote{The arrow direction is the attention direction, or the opposite direction of the information flow}, but also undesired interactions such as positive-to-negative $(\mathbf{P} \rightarrow \mathbf{N})$ and negative-to-negative $(\mathbf{N} \rightarrow \mathbf{N})$ (in which the value will cancel itself). These unintended interactions can distort the behavior of the model since the flipped signal influences more than just the image.

To address this, we introduce a duplication of the negative prompt. One copy remains unflipped and unscaled, denoted by $\mathbf{N}^{(0)}$, and the value (and only value) of the other is flipped and scaled, denoted by $V_{\mathbf{N}^{(1)}} = -\alpha \cdot V_{\mathbf{N}^{(1)}}$. The sequence becomes: $
[\mathbf{I},\ \mathbf{P},\ \mathbf{N}^{(0)},\ \mathbf{N}^{(1)}],
$ where $\mathbf{N}^{(1)}$ does not act as query in attention calculation. 


Additionally, inspired by \citet{wang_transpixeler:_2025}, where blocking some attention directions could improve quality, we apply attention masks to isolate the effect of $\textbf{N}^{(1)}$ to only $\textbf{I}$. Specifically, $\mathbf{N}^{(0)}$ is only allowed to attend to $\mathbf{I}$ and to itself, while $\mathbf{N}^{(1)}$ is only attended to by $\mathbf{I}$. Figure~\ref{fig:method} shows the attention mask. Since $\mathbf{N}^{(1)}$ does not act as a key or value in any attention query, it doesn't produce associated output. Instead, $\mathbf{N}^{(0)}$ serves as the negative prompt tokens passed to the subsequent MLP layer and into the next attention layer, where it will be flipped again. It acts as an information collector from images to collect unwanted elements and also keeps updating itself from attention to itself, matching the prompt updating in a positive prompt. 

This setup allows updates to the negative prompt based on attention from the image and from itself, and keeps the unflipped form active in the MLP path. It also prevents interference between positive and negative prompts and ensures that the flipped negative content affects only the intended image-to-negative attention path.

To preserve the high quality of generated images, we also applied attention bias ($-\beta$) to $\mathbf{I}\rightarrow\mathbf{N}^{(1)}$ (also shown in Figure~\ref{fig:method}) and we removed the padding tokens from the negative prompt. Details and pseudo-code of our method are in the Appendix. 

In our implementation, negative embeddings are duplicated once due to an implementation detail; however, this has a negligible impact on the final results after offsetting the scaling factor with 1 \footnote{\url{https://github.com/weathon/VSF/issues/19}}.

\section{Experiments}

\subsection{Dataset}
Following \citet{park_know_2025}, we use ChatGPT o3 \citep{o3} to generate pairs of prompts and negative prompts to construct our dataset NegGenBench. Unlike prior work, our prompts are intentionally more challenging: the negative prompt is typically related to the positive one, and as a critical component—e.g., the positive prompt of a bike could have a negative prompt of ``wheels". However, the positive prompt sometimes also uses a non-negation method to imply the item is missing, such as using terms like ``empty'' and ``exposed'' to make it more natural. Besides prompts, two questions are generated at the same time for later evaluation, one question asking if the image has the main object, either with or without the negative element, and the other one queries if the negative prompt element is missing. Prompts are generated in batches. 
There are 200 prompts generated, and we run them with 2 different seeds for the main results. 

\subsection{Baseline and Metrics}
We chose NAG \citep{chen_normalized_2025} and NASA \citep{nguyen_snoopi:_2024} as our baseline for few-step models. We also used a base model without negative guidance as a vanilla baseline, aiming to show the lower bound of the dataset. (i.e., how likely the positive prompt alone will help avoid negative concepts, if there is no negative guidance. This could happen either because the model is following the implication in the positive prompt, such as the word `missing', or simply by chance.) Because NASA's original source code was not publicly available at the time of writing, we reimplemented it based on NAG's codebase. Specifically, we replaced the guidance equation from NAG (Eq.~\ref{eq:nag1}) with NASA's equation (Eq.~\ref{eq:NASA}), removed normalization and blending, and enabled guidance when the scale is greater than 0 (instead of 1). Additionally, we compared our method in non-few-step models with CFG and used other models as external baselines (External baseline results are in Table~\ref{tab:external}, but the experiment details are in the Appendix). We included Flux Kontext \cite{labs_flux_knoest} as an image-editing baseline. We first generate an image using SD-3.5-Large-Turbo, and then edit the image with Flux Kontext using the prompt \texttt{Remove [negative prompt]}. Since NAG was focused on quality instead of negative prompt avoidance, we re-tuned its hyperparameter such that it has stronger negation following in trade-off of quality and positive prompt following. We name this variance as NAG Strong. Same for our VSF method, we provided two different variations with different hyperparameters, focusing on quality (VSF Quality) and negative prompt following (VSF Strong). Hyperparameter details are in the Appendix.



Following \citet{park_know_2025, wei_tiif-bench:_2025}, we used multimodal large language models (MLLM), specifically \texttt{llama-4-maverick-17b-128e-instruct-fp8}, to evaluate if the generated image follows the positive prompt and the negative prompt using the two questions generated during prompt generation. We did not use previous negation-aware CLIP-based work because they do not focus on missing an essential component, but simple meaning (e.g., a dog that is not on the grass). We did not use HPSv2 \citep{wu_human_2023} or ImageReward \citep{xu_imagereward:_2023} because they might give a low quality score for unusual objects (essential part being removed). Instead, MLLM is used to rate the image quality at the same time. At the end of our experiment, we also fine-tuned a Qwen-2.5-VL \citep{bai_qwen2.5-vl_2025} model using data we generated by VSF, NAG, and NASA for better negation understanding. More details about the metrics and comparison with human validation are in the Appendix.


\section{Results}
Quantitative results from using LLaMA as a judge evaluation are shown in Table~\ref{tab:res}, and qualitative results are shown in Figure~\ref{fig:res} in the Appendix. Human validation is shown in Table~\ref{tab:human} \footnote{The results here might be negligibly different than our code due to implementation differences. \url{https://github.com/weathon/VSF/issues/19}}. 
Automatic evaluation using the better negation-aware MLLM Qwen-2.5-VL is shown in Table~\ref{tab:res2} in the Appendix. Both the human validation and Qwen-2.5-VL results are aligned with our LLaMA evaluation relative ranking. It is important to highlight that the LLaMA assigns relatively generous quality scores; a score lower than 90 usually means the image already has degraded quality. Examples are in the Appendix Figure~\ref{fig:bad}.

\begin{figure}
    \centering
    \includegraphics[width=0.49\textwidth]{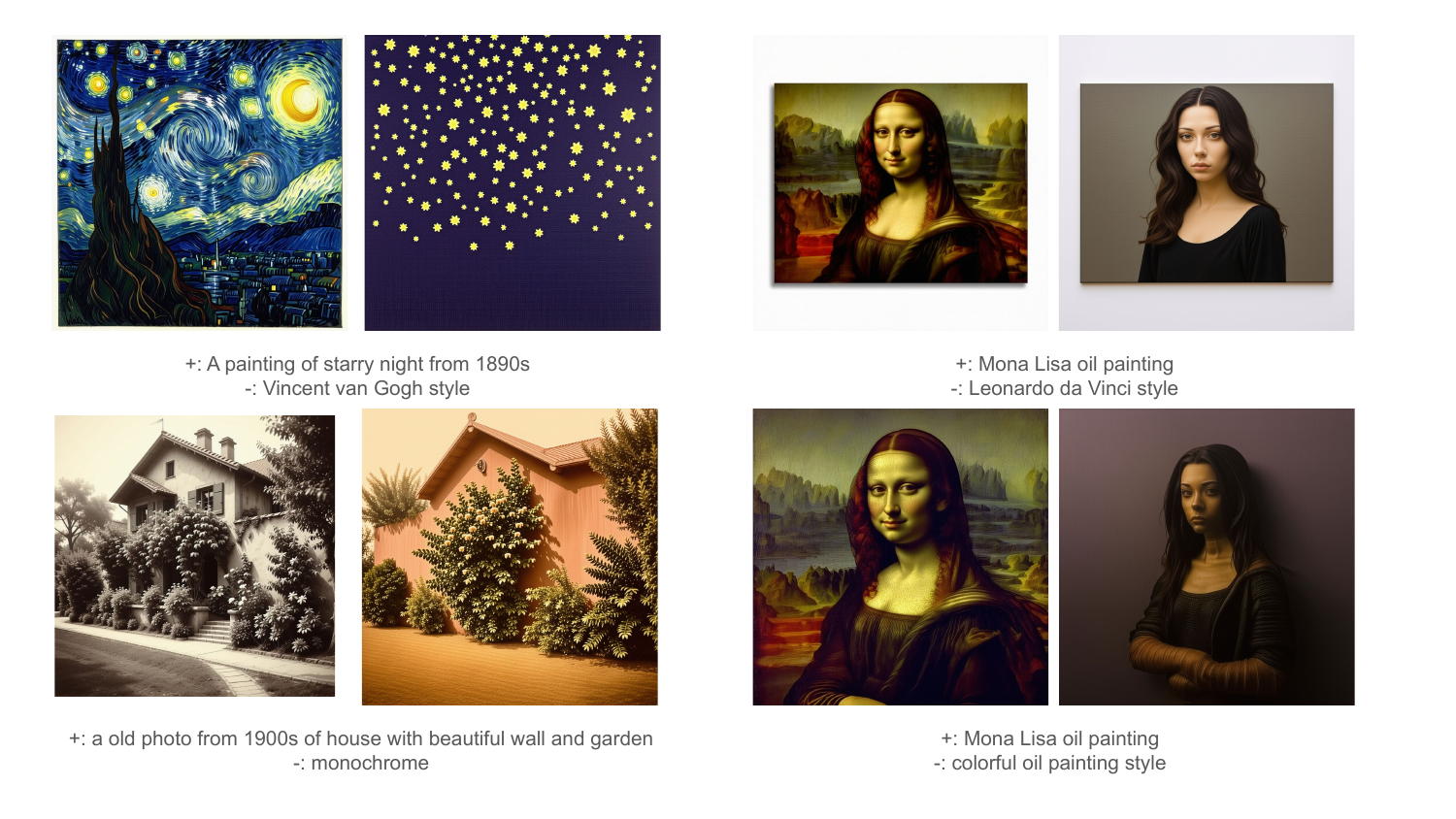}
    \raisebox{1em}{\includegraphics[width=0.49\textwidth]{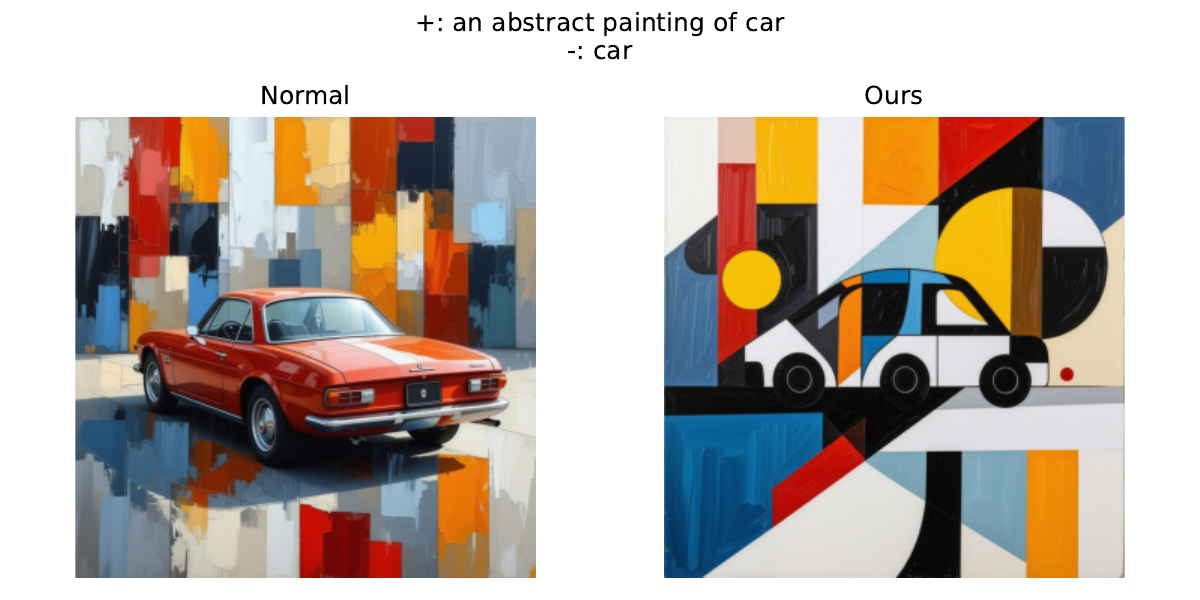}}
    
    \caption{(Left) Style Avoidance Tests, (Right) (Semi)-abstract art generated by mentioning the main object ``car'' in a negative prompt. The car is semi-canceled and thus still present but in an abstract form.}
    \label{fig:non_object}
    \vspace{-15pt}
\end{figure}




\begin{table}
    \centering
    \caption{External Baselines Comparsion}
    \small
    \begin{tabular}{c|cccc}
    \hline
         & Positive Score ($\uparrow$) & Negative Score ($\uparrow$) & Quality Score ($\uparrow$) &$\sim$Runtime ($\downarrow$)\\
     \hline
    \multicolumn{5}{l}{\textit{Open-weight Models}} \\
    \hline
    VSF Strong&  0.870&  \textbf{0.545}& 0.952 &\textbf{3s}\\
    VSF Quality& 0.980& 0.420& \textbf{0.986}&3s\\
    Generate+Edit     &  0.875&  0.488& 0.958 &55s\\
    Janus-4o    &  0.925&  0.225& 0.944 &20s\\
    Qwen-Image NP&  0.973&  0.190& 0.935&110s\\
    Qwen-Image Negation&  \textbf{0.990}&  0.100& 0.937&110s\\
    \hline
    \multicolumn{5}{l}{\textit{Closed-weight Models}} \\
    \hline
    GPT-4o     &  0.978&  \textbf{0.705}&  0.954&47s\\
    Nano Banana & \textbf{0.985} &0.498 & \textbf{0.980} & \textbf{14s} \\
     \hline
    \end{tabular}
    \label{tab:external}
    \vspace{-10pt}
\end{table}
Based on the quantitative results, VSF Strong shows a significantly higher negative score than other methods, while maintaining comparable or better quality scores. Our more conservative method, VSF Quality, still achieved the second-highest negative score, with the highest quality score. Both VSF Strong and VSF Quality even achieve a higher negative score than traditional CFG in non-few-step models, demonstrating a stronger ability to avoid negative elements, even relative to the established strong baseline. When compared with the external baseline, VSF also gets the highest performance in open source methods and only lags behind GPT-4o and achieves comparable performance with Nano Banana. 

We also tested concepts/style avoidance, as shown in Figure~\ref{fig:non_object} left. In the Starry Night example, VSF completely removed any signature elements in Vincent van Gogh's style, including the town, and generated a generic starry night image. Figure~\ref{fig:non_object} also illustrates how our method can produce abstract art, which is typically discouraged during a model's finetuning since reward models favor realism. This is achieved through using the same word as the main object for both positive and negative scores in VSF, as detailed in the Appendix. VSF also has the ability to generate ``anti-aesthetics'' (unconventional, including abstract) art. Details of these results are provided in Figure~\ref{fig:abstract2} and Figure~\ref{fig:ugy_art} (in the Appendix). 


\section{Discussion}
\subsection{Trade Off Curve}
\begin{figure}
    \centering
    \includegraphics[width=0.9\linewidth]{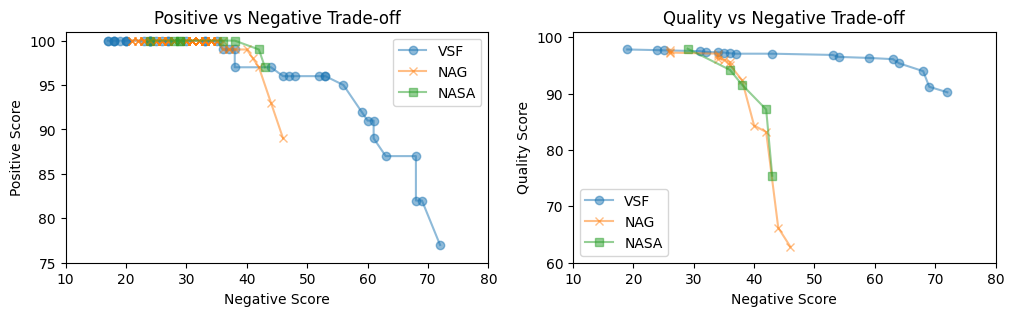}
    \includegraphics[width=0.9\linewidth]{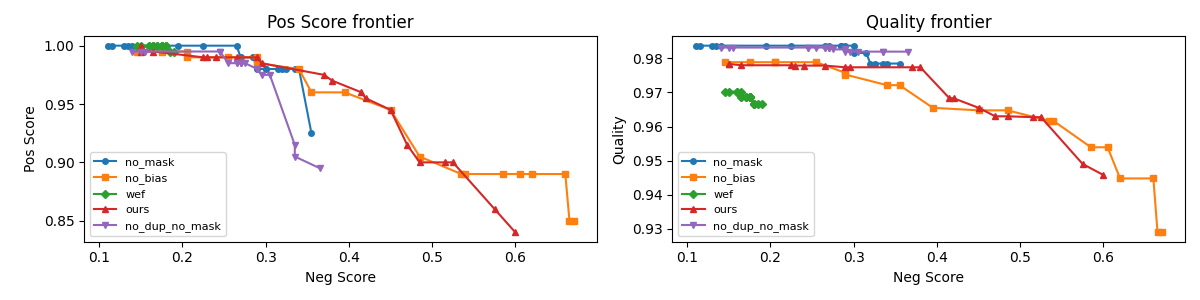}
    \caption{(Top) Trade-off plot of positive-negative score and quality-negative score. Both axies follows ``higher is better.'' (Down) Trade-off plot of the ablation study.}
    \label{fig:tradeoff}
    \vspace{-15pt}
\end{figure}

To systematically evaluate how effectively each model balanced positive prompt adherence, negative prompt adherence, and image quality, we conducted a hyperparameter sweep across each model. Specifically, we performed 66 runs for VSF and 287 runs for NAG, and 10 runs for NASA, with respect to their hyperparameter counts (2 for VSF, 3 for NAG, and 1 for NASA). A random sweep was executed besides for NASA, on which a single variable ``grid'' search is used, and evaluations were conducted using LLaMA, following the same criteria as previously described. Due to the large volume of runs, we limited our evaluation to the first 100 prompts with a single generation seed, potentially resulting in minor differences from earlier outcomes. Results are shown in Figure~\ref{fig:tradeoff} Left. 


From both plots, we observe that as the negative score increases, NAG and NASA both exhibit a significantly steeper and earlier decline compared to VSF in both positive and quality scores. In terms of positive score, VSF maintains scores above 90 even when the negative score rises to approximately 60. Regarding image quality, VSF similarly retains scores above 90 until a negative score of around 60, after which quality declines. In contrast, NAG and NASA both experience a sharp and early decline, with their quality score rapidly dropping to nearly 60 even before the negative score reaches 50. Keep in mind that a quality score under 90 means the image is already degraded, and if an image is rated 60, it is usually completely distorted. See Figure~\ref{fig:bad} in the Appendix for example. 

Additionally, VSF demonstrates a broader operational range in negative scores. When necessary, it can achieve negative scores exceeding 70 while still preserving acceptable positive prompt adherence and image quality. Conversely, NAG and NASA become unacceptable in quality at negative scores below 50, limiting their practical effectiveness.

\subsection{Attention Maps}
Since our proposed method performs adaptive steering based on a negative attention map, we visualize the attention maps generated during the diffusion process in Figure~\ref{fig:attn} in Appendix. Extracting the full attention maps is difficult because efficient implementations, such as FlashAttention, do not explicitly store these maps, and storing and computing them will require a large amount of memory. Therefore, we computed only the unnormalized attention values between the image tokens and negative prompt tokens. Figure~\ref{fig:attn} demonstrates that when the scale is set to 0, umbrellas appear, whereas setting the scale to 3 effectively avoids them. As indicated in the attention maps, image tokens corresponding to regions where umbrellas might exist (e.g., above human heads) exhibit higher attention toward the negative prompt tokens. Specifically, in steps 4 and 5, regions above the individuals on the left and right show strong negative attention, aligning with areas visually identified as umbrellas in $\alpha=0$ image. In the final image, these highlighted regions no longer contain umbrellas, confirming that our method effectively suppresses the presence of undesired objects at specific locations.

\subsection{Abliation Study}

To evaluate the effectiveness of each component of our approach, we conducted an ablation study using the following settings. For each setting, we scanned across scales for all 200 prompts using the same seed. Similar to before, we plotted the trade-off curve for each setting.

Rather than altering the attention values, we explored a simpler and more intuitive approach: flipping the text embedding prior to input into the DiT (Whole Embedding Flip, WEF). This is similar to applying the CFG on text embeddings studied in \citet{nguyen_snoopi:_2024}, but keeps the positive and negative tokens separated.  Specifically, the negative text embedding is scaled by $-\alpha$, concatenated with the positive prompt embedding in the sequence length dimension, and used as the prompt embedding for the DiT. We did not remove the padding for the negative prompt, as we found out that removing it causes the negative prompts to have no effect at all. 

We also tested our approach with no bias, no mask (but still duplication), and no duplication no mask. The trade-off plot is shown in Figure~\ref{fig:tradeoff} Right. The simpler and more intuitive WEF approach appears to have almost no effect at all. We hypothesize that this is because it is similar to flipping both the key and the value, causing regions most similar to the flipped key (i.e., least similar to the original negative prompt) to be pushed away, rather than pushing away regions most similar to the original negative prompt (i.e., unflipped key).
From the figure, we can see that the configurations without masking have a sharp positive score drop as the negative score increases. The WEF has a very limited range of negative scores. Our methods and the one without attention bias have similar results; this could be due to the MLLM not being sensitive enough to see the minor changes in quality.

Ablation study on hyperparameters is shown in the Appendix. 
\begin{table}
    \centering
    \caption{Positive scores (how well the model follows the positive prompts) and negative scores (how well the model avoids the negative prompts) of our model (VSF), NAG \citep{chen_normalized_2025}, and NAG with hyperparameter re-tuned (NAG Strong).}
    \begin{tabular}{c|ccc}
    \hline
         &  Positive Score ($\uparrow$)& Negative Score ($\uparrow$)&Quality Score ($\uparrow$)\\
         \hline
         VSF Strong&  0.870& \textbf{0.545}& 0.952\\
 VSF Quality& 0.980& 0.420&\textbf{0.986}\\
 NAG \citep{chen_normalized_2025}& 0.993& 0.220& 0.968\\
 NAG Strong & 0.975&0.320& 0.901\\
 NASA\citep{nguyen_snoopi:_2024}& 0.970& 0.380& 0.867\\
         None&  0.990& 0.195& 0.968\\
CFG \citep{ho_classifier-free_2022} (28 steps)& \textbf{1.000}& 0.300&0.956\\
         \hline
 
    \end{tabular}
    \label{tab:res} 
    \vspace{-15pt}
\end{table}

\begin{table}[]
    \centering
    \caption{Human Labelled Metric For 10 Selected Prompts with 2 Seeds}
    
·        \begin{tabular}{l|rrr}
        \hline
         & Positive Score ($\uparrow$)& Negative Score ($\uparrow$)& Quality Score ($\uparrow$)\\
        \hline
        NAG Strong & 0.950 & 0.250 & 0.675 \\
        NAG & \textbf{1.000}& 0.100 & \textbf{0.895}\\
        NASA & 0.950 & 0.150 & 0.685 \\
        VSF Quality & 0.900 & \textbf{0.550}& 0.823 \\
        \hline
        \end{tabular}
        
    \label{tab:human}
\end{table}

\section{Conclusion}
In this paper, we introduced VSF, a novel approach for enhancing negative prompt adherence in image and video generation models. Our method involves flipping the sign of attention values and duplicating negative prompts and attention masking, effectively suppressing unwanted content. Experimental results indicate that VSF significantly outperforms previous methods, NAG \citep{chen_normalized_2025}, NASA \citep{nguyen_snoopi:_2024}, and even CFG in terms of negative prompt adherence, with much lower trade-offs in overall quality and positive prompt following. We also showed that VSF can be applied to create more creative (by style avoidance, abstract images, and anti-aesthetics styles) images. VSF also only has one main hyperparameter and one minor hyperparameter, making it easier to tune them in downstream tasks. Future work directions are discussed in the Appendix.


\section*{Acknowledgements}
This work was supported by the NFRF under grant GR024801 and the CFI under grant GR024473. The authors acknowledge Weathon Software (\url{https://weasoft.com}) and Lambda, Inc. (\url{https://lambda.ai/}) for providing computing resources via Google Colab and Lambda Cloud, respectively.


Following the completion of this research and paper submission, we became aware of NegPiP (\url{https://github.com/hako-mikan/sd-webui-negpip/}), a community project that utilizes an approach similar to ours. While this work was developed entirely independently from NegPiP, we recognize its contribution and suggest it as a relevant resource for readers.


\bibliography{refs}
\bibliographystyle{iclr2025_conference}

\appendix
\addcontentsline{toc}{section}{Appendix} 
\newpage  
\part{Appendix} 
\parttoc
\section*{Appendix}
\section{Negation in Vision Language Models}
\begin{figure*}
    \centering
    \includegraphics[width=\linewidth]{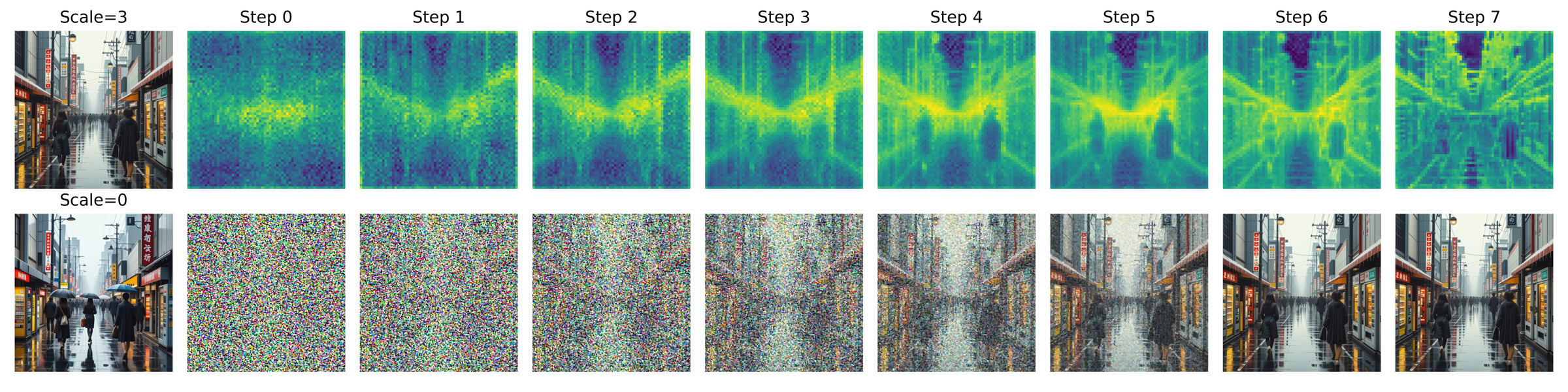}
    \caption{Attention maps and intermediate images during the diffusion process. The leftmost column shows the final generated image (top) and an image generated without applying VSF scaling ($\alpha=0$, bottom). The top row on the right side displays the unnormalized attention values between image tokens and negative prompt tokens, while the bottom row shows the corresponding intermediate images at each timestep. The negative prompt is ``umbrella.''}
    \label{fig:attn}
\end{figure*}
\begin{figure}
    \centering
    \includegraphics[width=0.5\linewidth]{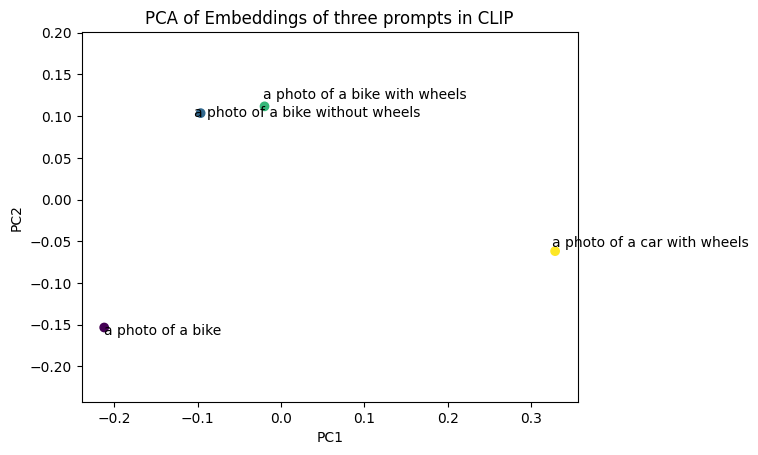}
    \caption{PCA plot of the 3 different prompts with a negation prompt.}
    \label{fig:PCA}
\end{figure}

Much previous work has shown that existing vision language models (VLM) struggle to understand negation \citep{yuksekgonul_when_2023, singh_learn_2024, alhamoud_vision-language_2025, park_know_2025}. In classification tasks, the model cannot correctly understand text with negation in it, e.g. ``a dog running" vs ``a dog not running" might have very close embeddings, even though they are opposite. In our test using a CLIP-ViT-Base-32, the cosine similarity between ``a dog running'' and ``a dog not running'' is 0.9243, where the similarity between ``a dog'' and ``a dog running'' is only 0.8710. In Figure~\ref{fig:PCA}, we show 4 prompts ``a photo of a bike", ``a photo of a bike without wheels", ``a photo of a bike with wheels", and ``a photo of a car with wheels". We can see that the bike with wheels and the bike without wheels have extremely close embeddings.
This problem has been introduced into text-to-image generation tasks, making it hard for the model to generate images without certain concepts (examples in Figure 1 of \citet{singh_learning_2025} and Figure 5 of \citet{park_know_2025}). Thus, classifier-free guidance (CFG) was used to introduce a negative prompt to the image generation process. More details in the next section. 
Several studies have attempted to tackle this issue by employing alternative training strategies, such as incorporating harder samples in the training data designed for negation tasks \citep{yuksekgonul_when_2023, singh_learn_2024, singh_learning_2025, alhamoud_vision-language_2025, park_know_2025}. Some of these methods have shown improvements in image generation tasks. For instance, \citet{park_know_2025} reported gains in Neg Score—measuring whether the model retains the primary subject while correctly omitting the negated object—for both SD-1.4 and SDXL-1.0, by replacing the default CLIP encoder with their NegationCLIP on their dataset, without additional T2I training. 

These methods generally require re-training the text encoder (usually a CLIP-like model) with contrastive learning, which poses challenges for models that do not use contrastively pre-trained encoders, such as T5 \citep{raffel_exploring_2023} in Stable Diffusion 3 \citep{sd3, esser_scaling_2024} and Flux \citep{noauthor_black-forest-labs/flux.1-schnell_2025}. Moreover, each model using a different text encoder would require a separate, dedicated adaptation. Additionally, even if the text encoder understands the negation, the diffusion model might still fail to avoid certain items because of their strong association. 

\section{Proof That Our Method is the Same as Token-Weighted Subtraction}
In this section, we prove that our method $Z^{VSF}$ is equivalent to token-weighted subtraction, denoted $Z^{W}$. 
\begin{proof}
We define
\[
A^+ = \exp(\frac{Q(K^+)^T}{\sqrt{d}}), \quad A^- = \exp(\frac{Q(K^-)^T}{\sqrt{d}}).
\]

Then
\[
W = \frac{\sum A^+}{\sum A^+ + \sum A^-}.
\]

Substituting into the expression for $Z^W$:
\[
Z^{W} = W \cdot \sigma(Q(K^+)^T)V^+ - (1 - W) \cdot \alpha \cdot \sigma(Q(K^-)^T)V^-,
\]
and using the softmax definitions
\[
\sigma(Q(K^+)^T) = \frac{A^+}{\sum A^+}, \quad \sigma(Q(K^-)^T) = \frac{A^-}{\sum A^-},
\]
we obtain
\[
\]

Canceling the sums:
\[
Z^{W} = \frac{A^+}{\sum A^+ + \sum A^-} V^+ - \frac{\alpha A^-}{\sum A^+ + \sum A^-} V^-.
\]

This matches
\[
Z^{VSF} = \sigma(Q(K^+ \oplus K^-)^T)(V^+ \oplus -\alpha V^-),
\]
since
\[
\sigma(Q(K^+ \oplus K^-)^T) = \frac{A^+\oplus A^-}{\sum A^+ + \sum A^-},
\]
and therefore
\[
Z^{VSF} = \frac{A^+}{\sum A^+ + \sum A^-} V^+ + \frac{A^-}{\sum A^+ + \sum A^-} (-\alpha V^-) = \frac{A^+}{\sum A^+ + \sum A^-} V^+ - \frac{\alpha A^-}{\sum A^+ + \sum A^-} V^-.
\]

Hence, $Z^W = Z^{VSF}$.
\end{proof}

\section{Attention Bias and Padding Removal}
We observe that even when the scaling factor $\alpha = 0$, including the negative prompt in the sequence still sometimes reduces image quality. This could be because the negative prompt ``distracts'' the image tokens' attention from the image tokens or positive prompts. To mitigate this effect, we introduce a negative bias $-\beta$ into the attention $\mathbf{I} \rightarrow \mathbf{N}^{(1)}$, thereby reducing the influence of the negative prompt.

In most models from Huggingface Diffusers \citep{von_Platen_Diffusers_State-of-the-art_diffusion}, padding tokens in the text input are typically not masked during attention. This is likely because the models have learned to ignore padding, and masking them would add unnecessary overhead (due to some attention implementations like FlashAttention-2 \citep{dao_flashattention-2:_2023} that do not support arbitrary masking). However, when we invert the sign of the padding tokens, it degrades output quality significantly. This could be because, although these tokens carry no semantic meaning, the sign-flipping introduces unseen states into the attention mechanism. To mitigate this, we remove padding tokens from the negative prompt embeddings. For the positive prompt, we retain padding tokens, as they do not introduce novel tokens and can improve generation quality. This aligns with training conditions and may allow the model to use padding positions as registers for auxiliary information.

\section{Details About the Metrics}
\begin{figure}
    \centering
    \includegraphics[width=0.2\linewidth]{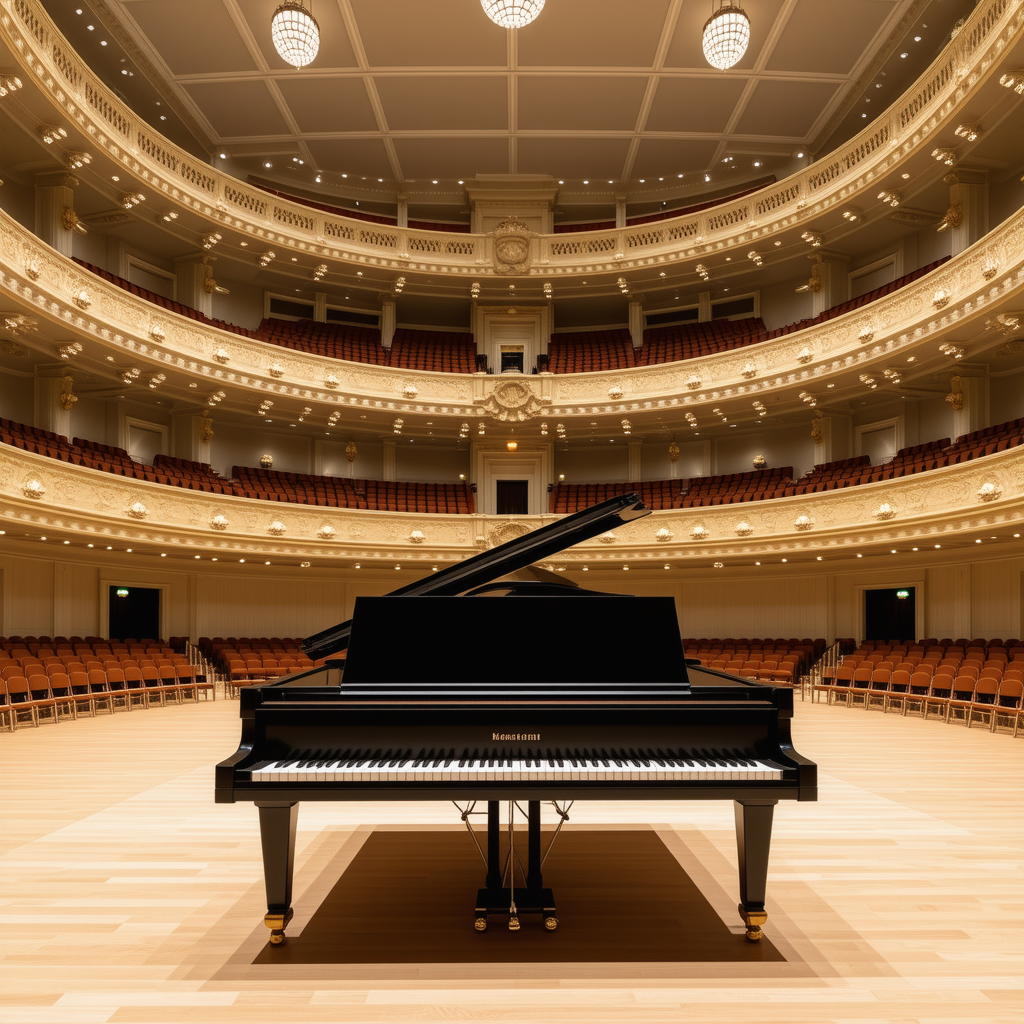}
    \includegraphics[width=0.2\linewidth]{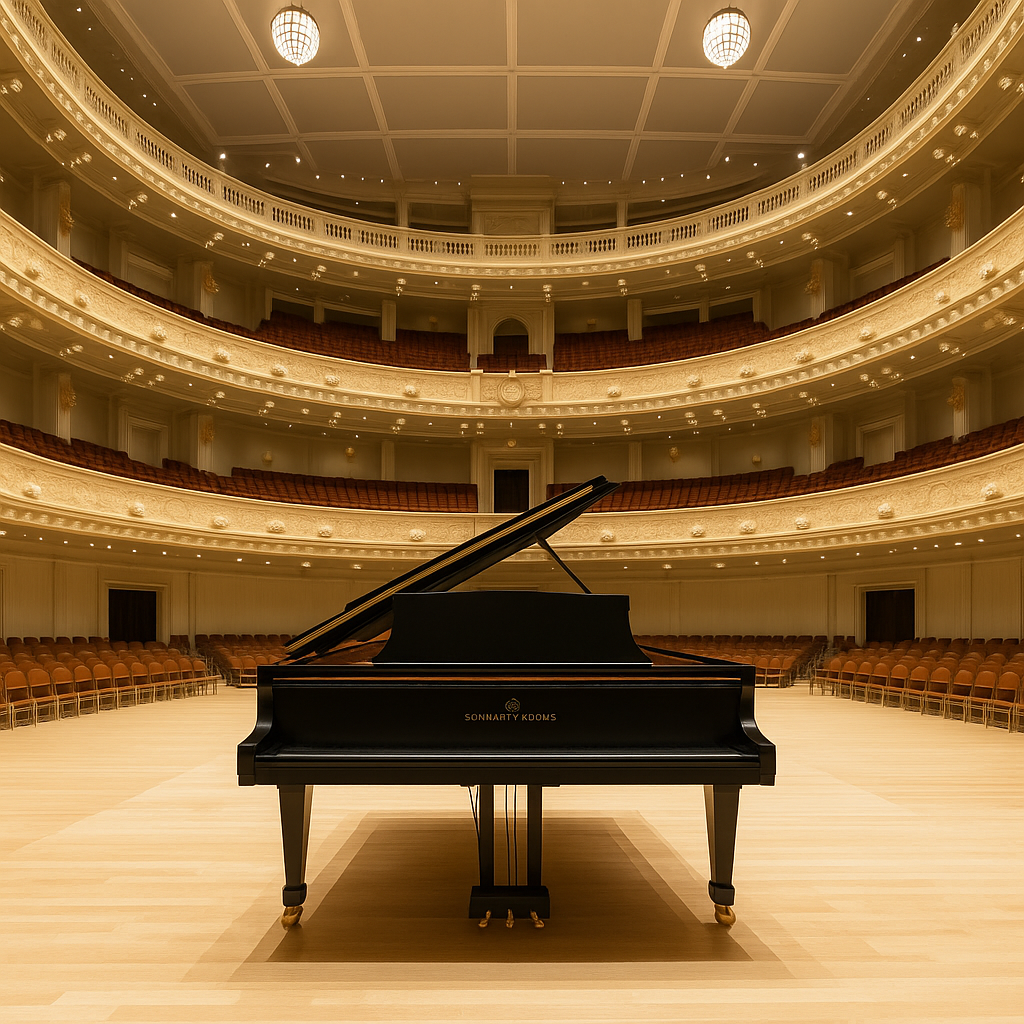}
    
    \caption{The left image is the original image, and the right image is generated by GPT-4o, where piano keys are removed. When scored using HPSv2, the left image got a score of 0.330 while the right image got a score of 0.319 using prompt of ``A grand piano dominates an empty concert hall, a smooth ebony board stretching across the front.'' and the left get a score of 0.330 and the right got a score of 0.324 if we mention ``no keys'' in the promps.}
    \label{fig:no_keys}
\end{figure}
To evaluate the scores of the generated images, we used LLaMA 4 Maverick, which has a very high image reasoning MMMU \citep{mmmu} score, higher than Gemma 3 and even GPT-4o. We avoided using the same model (o3) for both evaluation and generation for cost control and to avoid bias within a model. We did not evaluate the quality of the generated images using models like ImageReward \citep{xu_imagereward:_2023} or HPSv2 \citep{wu_human_2023} as in NASA or NAG, as current quality or human preference assessment models do not account for negative prompts; traditional methods usually aim for real-world generations \citep{ye_echo-4o:_2025}. Removing a key element from the positive prompt (e.g., removing the roof from a house) is likely to reduce perceived quality, since the result deviates from what is considered ``normal,'' even though that is the intended outcome. An example is shown in Figure~\ref{fig:no_keys} where removing the keys from the piano results in a much lower quality score, even though other parts of the image look the same. Both ImageReward and HPSv2 are built on top of image-text alignment models (CLIP \citep{radford_learning_2021} or BLIP \citep{li_blip:_2022}), which will likely lead to a decreased score when the main object is missing a critical part. Thus, we also let the MLLM rate the image quality from 0-1 for each image and told it to ignore the abnormality of following the negative prompt. 
We did observe that MLLM can make mistakes, especially when there is ambiguity or when the unwanted element is hard to see. However, we believe that in general, under 400 images, the mistakes are minor.  To compensate for this, we also did a human evaluation and a more negotiation-aware MLLM evaluation to cross-validate the LLaMA evaluation. Due to model provider stability issues, we used different MLLM providers for different portions of the experiment for the same model under the same config; this could have some limited impact on the stability of metrics.
\begin{figure}
\begin{lstlisting}[
    language=Python,
    caption={Pseudocode implementation of the Value Sign Flip (VSF) attention process},
    basicstyle={\ttfamily\tiny}
]
# prep for embeddings
pos_embeds = get_embed(prompt)
neg_embeds = get_embed(neg_prompt, padding=False)
pos_len, neg_len, img_len = pos_embeds.shape[1], neg_embeds.shape[1], IMG_LEN

# concat positive and negative prompts (N0)
prompt_embeds = torch.cat([pos_embeds, neg_embeds], dim=1)

# prep for attention mask and bias (N1 never acts as query)
total_len = img_len + prompt_embeds.shape[1]
attn_mask = torch.zeros((1, total_len, total_len + neg_len))

# block P and N0 from attending to N1
attn_mask[:,-(pos_len+neg_len):,-neg_len:] = -torch.inf

# block image and P from attending to N0
attn_mask[:,:-neg_len,-(2*neg_len):-neg_len] = -torch.inf

# block N0 and N1 from attending to P
attn_mask[:,-neg_len:,img_len:img_len+pos_len] = -torch.inf

# bias image->N1 connections
attn_mask[:,:img_len,-neg_len:] -= offset


class VSFAttnProcessor(AttnProcessor):
    def __init__(self, attn_mask, neg_prompt_length):
        self.attn_mask = attn_mask
        self.neg_prompt_length = neg_prompt_length

    def forward(self, hidden_states, encoder_hidden_states, attention_mask):
        # get qkv projection for image tokens
        q = self.get_q(hidden_states)
        k = self.get_k(encoder_hidden_states)
        v = self.get_v(encoder_hidden_states)

        # get qkv projection for encoder tokens

        q_enc = self.get_q_encoder(encoder_hidden_states)
        k_enc = self.get_k_encoder(encoder_hidden_states)
        v_enc = self.get_v_encoder(encoder_hidden_states)

        query = torch.cat([q, q_enc], dim=2)

        # append P, N0 (in k_enc and v_enc) and N1 (the last portion of k_enc and v_enc) at the end
        k = torch.cat([k, k_enc, k_enc[:,:,-self.neg_prompt_length:]], dim=2)
        v = torch.cat([v, v_enc, v_enc[:,:,-self.neg_prompt_length:]], dim=2)

        # sign-flip values of N1
        v[:,:,-self.neg_prompt_length:] *= -scale

        hidden_states = F.scaled_dot_product_attention(
            query, k, v,
            dropout_p=0.0, is_causal=False,
            attn_mask=self.attn_mask.to(query.dtype)
        )
        hidden_states = hidden_states.transpose(1, 2).reshape(batch_size, -1, attn.heads * head_dim)
        return self.out_proj(hidden_states)


for block in model.transformer.blocks:
    block.attn1.processor = VSFAttnProcessor(attn_mask, neg_len)

# diffusion process continues
\end{lstlisting}
\end{figure}

\section{Hyper-parameter Tunning}
Although NAG \citep{chen_normalized_2025} also targeted negative concept avoidance, its primary focus was on its effects on improving generation quality (using words like ``blurry" or ``low quality'' as a negative prompt). We believe the hyperparameters reported in their work were tuned with an emphasis on quality rather than negation handling. Therefore, we re-tuned their hyperparameters moderately and manually on guidance scale ($\phi$), blending factor ($\alpha$), and normalization factor ($\tau$). We will report experimental results on both original NAG (noted as NAG) and the improved hyperparameter version (noted as NAG Strong). The final hyperparameters used are $\phi=11, \alpha=0.5, \tau=5$ for NAG Strong and $\phi=4, \alpha=0.125, \tau=2.5$ for original NAG.
This pushes the NAG to the edge of acceptable visual quality.

Similarly, for our VSF, we used two set of hyperparameters, VSF Quality ($\alpha=3.3,\beta=0.2$) maintained high quality and positive prompt alignment, while VSF Strong ($\alpha=3.8,\beta=0.2$) pushes it to the limit, reaching higher negative prompt alignment in trade of positive and quality score. 
\section{Human Validation}
\begin{figure}
    \centering
    \includegraphics[width=0.5\linewidth]{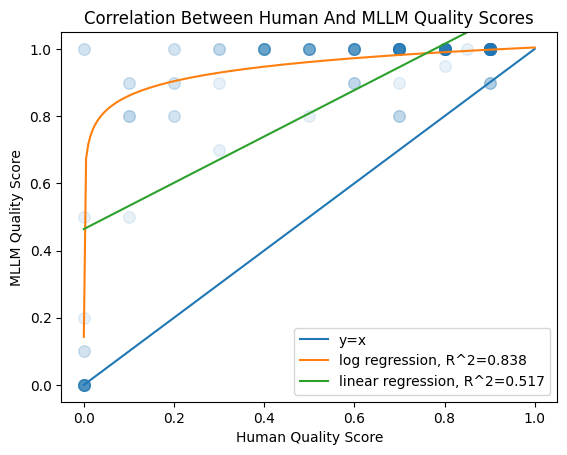}
    \caption{Correlation Between the Human-rated quality score and MLLM-rated quality score}
    \label{fig:human_mllm_score}
\end{figure}

To verify the results of the MLLM evaluation, we selected 10 prompts (with 2 seeds each) for VSF, NASA, NAG, and NAG Strong and manually labeled them with positive, negative, and quality scores. The human-labeled results are presented in Table~\ref{tab:human}. We validated MLLM performance by computing the binary F1 score and accuracy between MLLM outputs and human annotations. Cohen’s kappa was not applied due to the highly imbalanced class distribution. The reliability metrics are summarized in Table~\ref{tab:rel}. We observed that quality ratings from MLLM and human labels were uncorrelated in high-quality regions. To investigate this further, we evaluated quality scores over a broader set of conditions. With a large sample size, we found that correlation emerges in a wider range: when scores are close to 1, small fluctuations carry little meaning, but substantially lower scores (e.g., $<0.9$) may indicate degraded quality. The correlation is shown in Figure~\ref{fig:human_mllm_score}. This supports the observation in Figure~\ref{fig:bad}, where MLLM tends to overestimate quality. From the scatter plot and regression, we can see that the MLLM score is usually higher than the human score, and although they are not linearly correlated, they are monotonically correlated.

\begin{table}[]
    \small
    \centering
    \small
    \caption{Reliability Metric For Human And MLLM Evaluation Results}
    
        \begin{tabular}{l|rr}
        \hline
         & Negative Score & Positive Score \\ \hline
        F1 & 0.667 & 0.974 \\
        Accuracy & 0.850 & 0.950 \\
        \hline
        
        \end{tabular}
    \label{tab:rel}
\end{table}

\section{Adapting to Other DiT Models}
In this paper, we primarily use SD-3.5 \citep{sd3} due to simplicity and elegant architecture. However, our method can theoretically be adapted to any transformer-based diffusion or flow-matching model. To demonstrate this adaptability, we implemented our method on Wan 2.1 with CausVid LoRA \citep{yin_slow_2025, noauthor_kijai/wanvideo_comfy_nodate} and Flux Schnell \citep{noauthor_black-forest-labs/flux.1-schnell_2025}.

For Wan 2.1, which uses cross-attention between image and text, duplication and masking are unnecessary and not used. Because our approach does not perform extrapolation and solely provides negative guidance, it cannot enhance overall quality significantly or replace CFG sampling in non-disstilled models, making it incompatible with the original Wan 2.1 model. Instead, we utilize CausVid \citep{yin_slow_2025}, which enables Wan to function effectively without classifier-free guidance in few-step settings. Specifically, we used a LoRA distilled from the original CausVid that can be directly applied on top of Wan 2.1 \citep{noauthor_kijai/wanvideo_comfy_nodate}. For qualitative results from Wan, please see the appendix. 

We also tested our method on Flux Schnell \citep{noauthor_black-forest-labs/flux.1-schnell_2025}. However, due to the model likely being trained to associate items with their associated items that often appear together strongly, we need to make some modifications. Before the negative prompt was fed into the model, we did a CFG-like extrapolation on the negative prompt, with a mean padding embedding as a null condition. This follows the implementation of the Compel package. Noted as:
\begin{equation}
    p^- = p^- + \lambda \cdot(p^- - p^\emptyset),
\end{equation} we used $\lambda=8$ in this case. Quantitative results are shown in Table~\ref{tab:flux}. We can observe that even without any negative guidance, the Flux Schnell model can slightly better avoid the items solely based on the positive prompts (since the positive prompts implied the item is missing using terms like ``empty''), but with the help of VSF, it further increases the negative score without compromising the positive and quality score. 

\begin{table}[]
\small
    \centering
    \caption{Comparsion of Flux Schnell VSF and  Original Schnell \citep{noauthor_black-forest-labs/flux.1-schnell_2025}}
    
    \begin{tabular}{c|ccc}
    \hline
     Method & Positive Score & Negative Score & Quality Score \\ \hline
     Flux Schnell& 1.00& 0.22& 0.99\\
 Flux Schnell VSF& 0.97& 0.41&0.99\\
 \hline
    \end{tabular}
    \label{tab:flux}
\end{table}

\section{Computational Cost}
Since our method does not require two passes through the entire model (as in CFG) or the attention module (as in NAG or NASA), and only slightly increases the sequence length ($<0.2\%$), its theoretical computational cost is significantly lower, close to that of a single pass. However, due to implementation limitations (specifically, FlashAttention-2's lack of support for arbitrary attention masking), the actual runtime of our method is higher than the original single-pass MM-DiT models, and similar to NAG or NASA, but still lower than CFG. 


\begin{table}
    \centering
    \small
    \caption{The computation cost of each model. Time is measured in total runtime per sample, and VRAM is the peak RAM during the 25 samples generation. Since VSF Wan does not require a mask, and it is only used for bias, we also tested it without the bias. The SD3.5 model used is SD-3.5-Large-Turb,o and the Wan model used is Wan-2.1-T2V-1.3B.} 
    
    \begin{tabular}{l|cccc}
        \hline
         & \multicolumn{2}{c}{Wan} & \multicolumn{2}{c}{SD3.5} \\
         & Time & VRAM & Time & VRAM \\
        \hline
        Baseline & 23.10s & 22.05GB & 2.14s & 28.49GB \\
 NASA& -& -& 2.89s&28.50GB \\
NAG & 25.58s & 22.06GB & 2.98s & 28.50GB \\
CFG (Theroatical) & 46.20s & - & 4.28s & - \\
VSF & 22.70s & 23.05GB & 3.00s & 28.53GB \\
VSF (No mask/bias) & 22.70s & 22.05GB & - & - \\

        \hline
    \end{tabular}
    \label{tab:compute}
\end{table}

To accurately measure the computational cost, we evaluate the runtime of 25 identical prompts under four settings: no guidance, NAG, NASA, and our proposed guidance, VSF, and then report the average runtime and peak memory usage for each setting. We also reported the theoretical CFG time as double the one without guidance. To avoid GPU thermal throttling affecting the results, we pause for at least 5 minutes between each set of tests. The tests are done on NVIDIA A100 40GB on Google Colab, as this is the most accessible option for high-end GPUs for users. Stable-Diffusion-3.5-Large-Turbo is generated in 8 steps for 1024x1024 resolution, Wan is generated in 8 steps with 480x832 resolution, and 81 frames. The results are shown in Table~\ref{tab:compute}.

From the table, VSF requires marginally more time and memory than NAG in SD3.5, while they are both significantly faster than theoretical CFG time, which would be twice the baseline. In Wan, VSF outperforms NAG and is even slightly better than the baseline (likely due to nature variation or noise) in terms of compute time, though it consumes 1GB more memory, likely due to the attention bias being stored. Since this bias is optional, we tested VSF Wan's performance with it removed, which results in an improvement in VRAM usage such that it uses the same amount of VRAM as baseline and NAG, and no change in runtime. 

\begin{figure*}
    \centering
    \includegraphics[width=\linewidth]{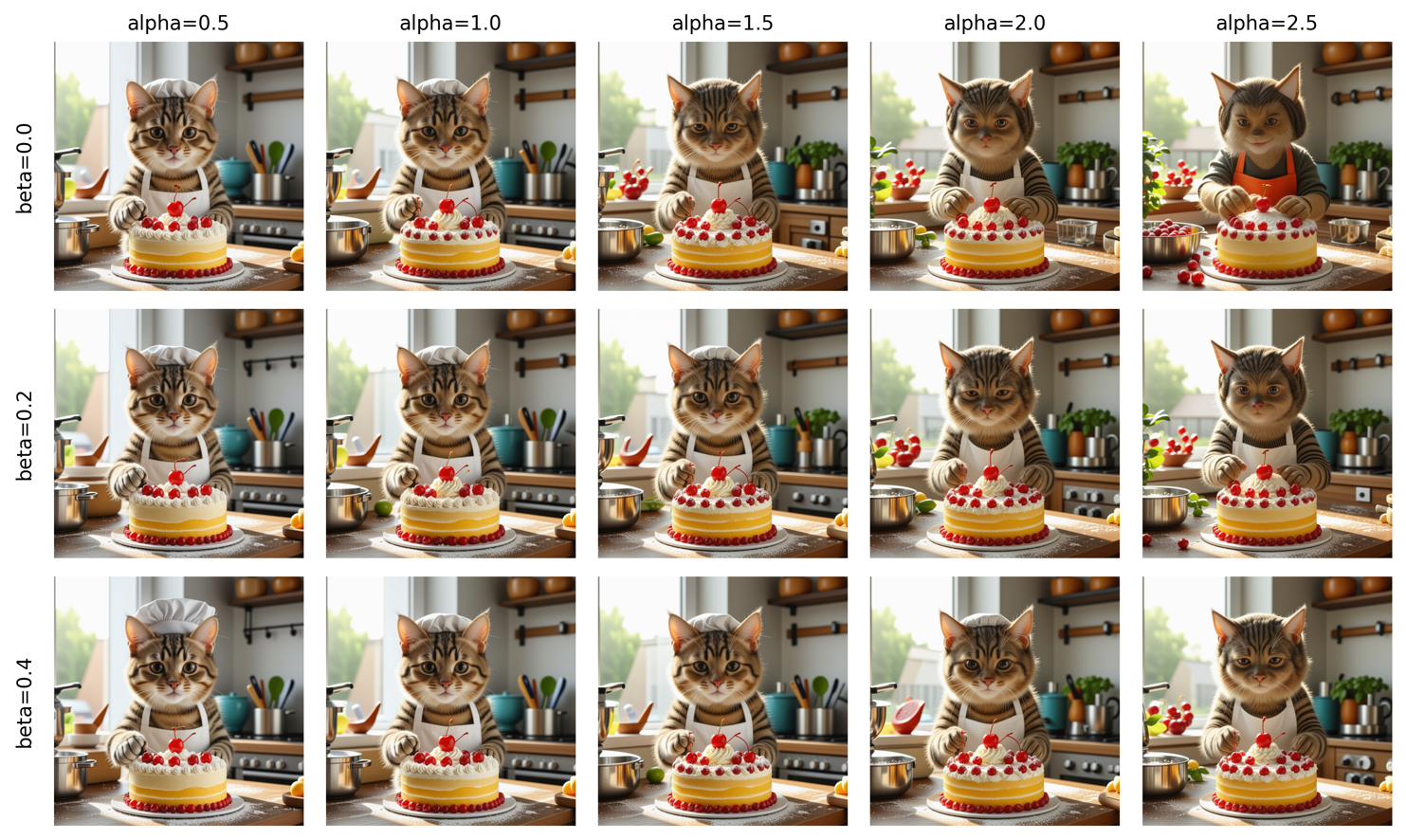}
    \caption{Effects of guidance scale ($\alpha$) and attention bias ($\beta$) in image generation. Positive prompt is ``a cat making a cake in the kitchen, the cat is wearing a chef's apron..." and negative prompt is ``chef hat."}
    \label{fig:effects} 
\end{figure*}

\begin{figure}
    \centering
    \includegraphics[width=0.3\linewidth]{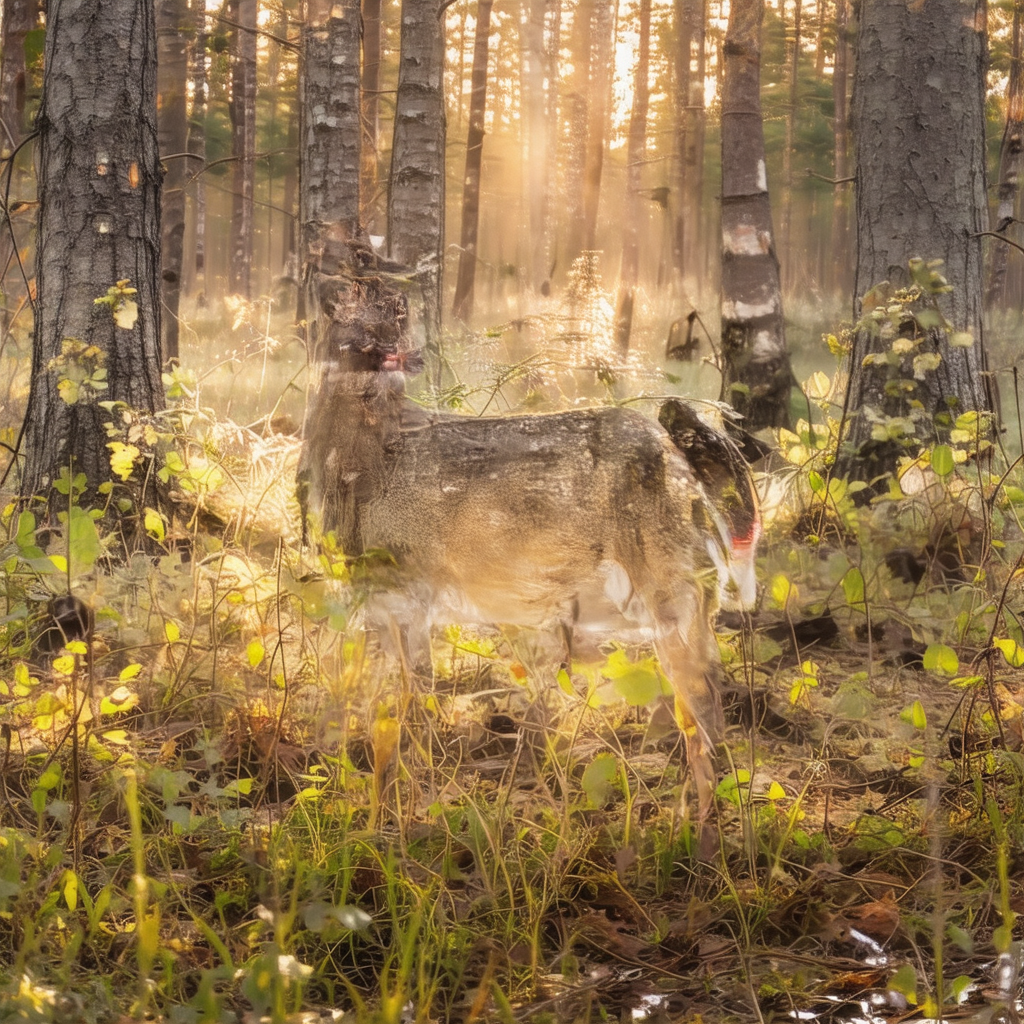}
    \includegraphics[width=0.3\linewidth]{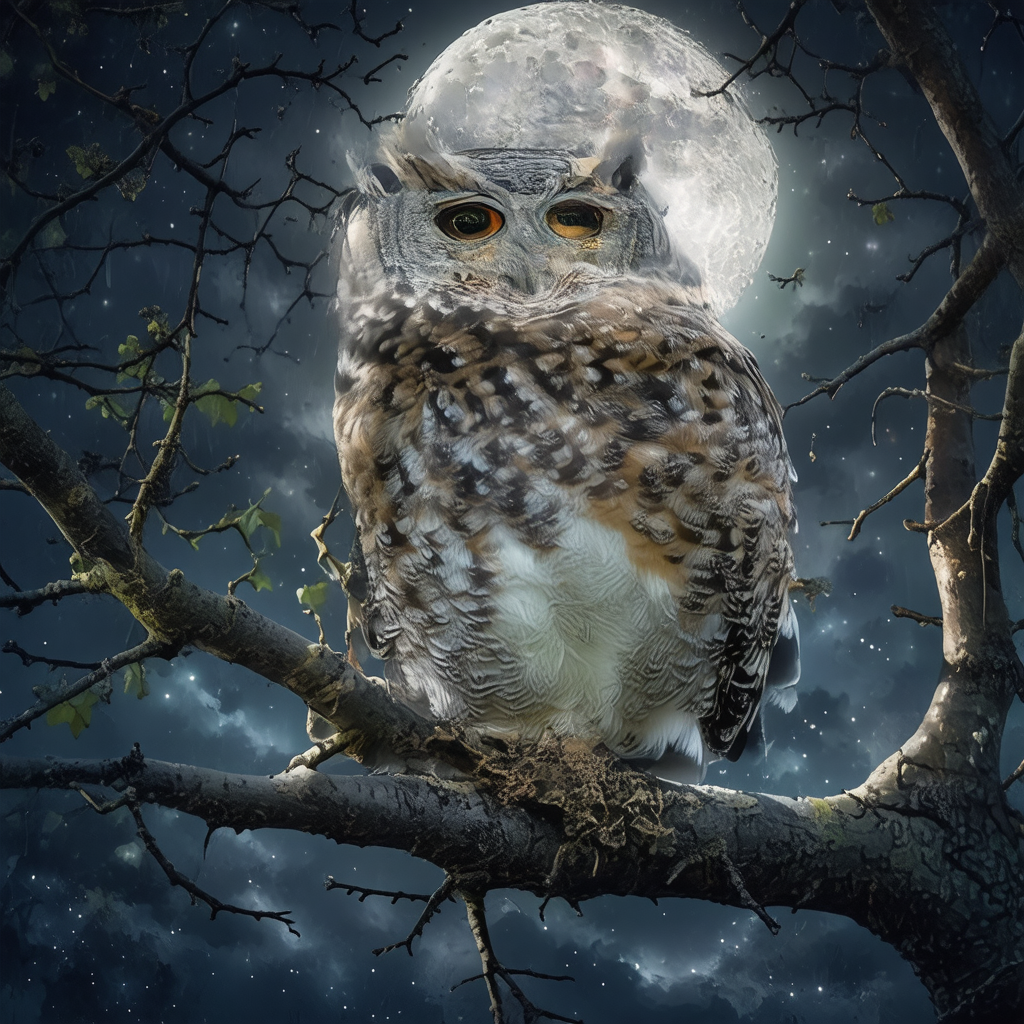}
    \includegraphics[width=0.3\linewidth]{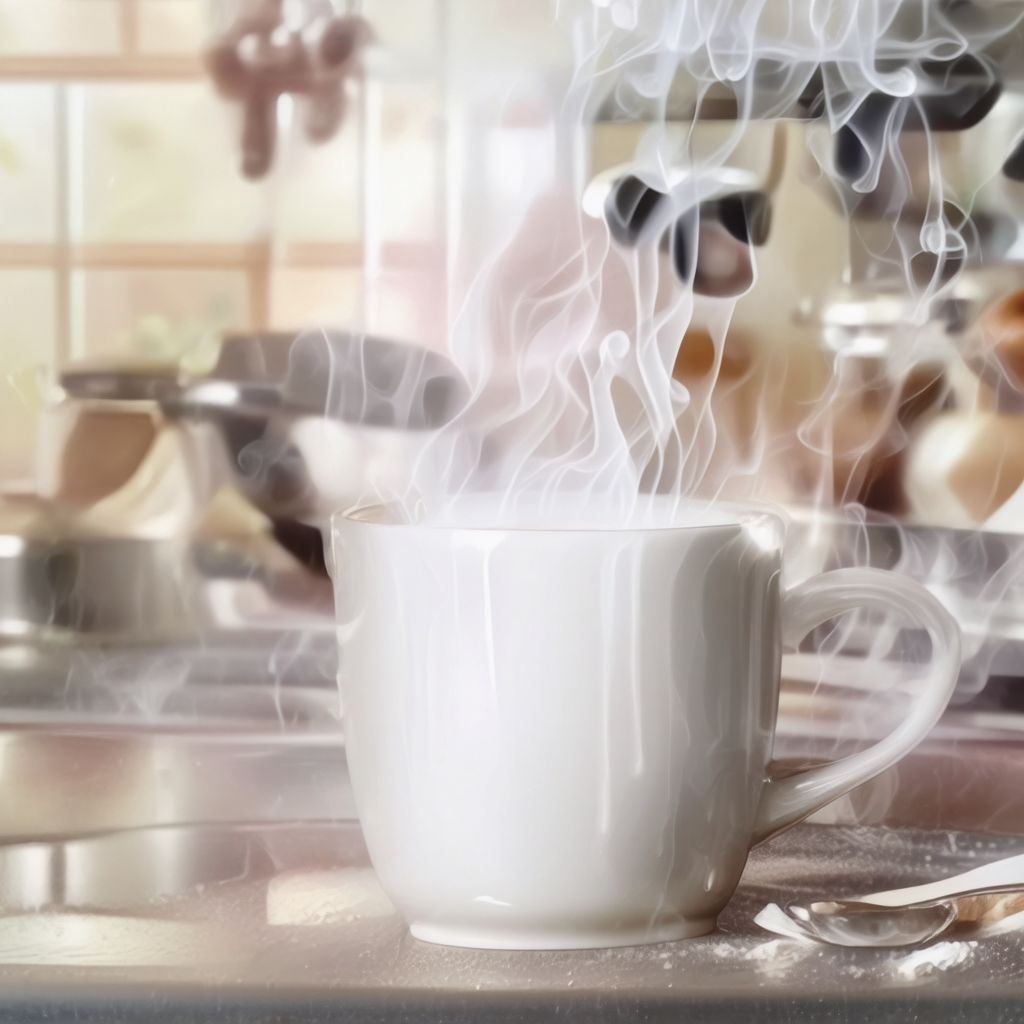}
    \caption{An example of a completely distorted image gets a relatively high quality score. The left one has a score of 70, the middle one has a score of 90, and the right one is a slightly distorted image, but still rated for 100. }
    \label{fig:bad}
\end{figure}

\section{External Baselines}
In addition to other guidance methods applied to SD-3.5-turbo, we evaluated several external baselines. The first baseline employs a generate-then-edit approach, loosely inspired by Generate-Plan-Edit (GraPE) \citep{goswami_grape:_2025}, but omitting the planning stage as our goal is straightforward (removing unwanted elements). Specifically, we first generated images using SD-3.5-Large-Turbo without a negative prompt, and subsequently edited out the unwanted elements automatically using Flux Knoest \citep{labs_flux_knoest}, an image editing model, using prompt \texttt{Remove [negative prompt].} 

The second baseline utilizes GPT-4o's native image generation capability. GPT-4o has demonstrated strong prompt-following performance\citep{wei_tiif-bench:_2025}, including in handling negation tasks. As GPT-4o lacks explicit negative prompt functionality, we formatted prompts as \texttt{[Positive prompt], but with no [negative prompt]}. Since our focus is on evaluating negation rather than image quality, we adopted the ``low" generation setting. Besides GPT-4o, we also added the newly released Nano Banana from Google. It is also a language model-based image generation model and has received a good reputation in the image generation community. 

The third baseline we included is Janus-4o \citep{chen_sharegpt-4o-image:_2025}, a model distilled from GPT-4o onto the Janus-Pro base architecture \citep{chen_janus-pro:_2025}. Given GPT-4o’s strong prompt-following performance, we anticipated competitive results from Janus-4o. We provided negative prompts directly as negations within the positive prompts, same as GPT-4o.

Finally, we tested Qwen-Image \citep{wu_qwen-image_2025} using two configurations: one employing separate positive-negative prompt pairs using CFG (labeled as Qwen-Image NP), and another embedding negative prompts as negations within the positive prompt itself (labeled as Qwen-Image Negation), while still using CFG with an empty negative prompt. Qwen is run under DFloat-11.

All measure time is measured on Google Colab 40GB A100 GPU, and for Qwen-Image and Generate+Edit, model CPU offloading is enabled. 

The results are presented in Table~\ref{tab:external} in the main text. The table indicates that VSF Strong achieves the second-best negative score, only behind GPT-4o, while also demonstrating a significantly faster runtime compared to all other methods, outperforming even the generate-then-edit pipeline. The GPT-4o distilled model, Janus-4o, has an unexpectedly low negative score, which could be because they did not have enough negation-included prompts in the distillation data. The VSF Quality had a lower negative score compared to Generate+Edit, while having a much higher positive and quality score, and shorter runtime. 

\section{SD3.5-Large-Turbo Qualitative Results}
\begin{figure}
    \centering
    \includegraphics[width=0.8\linewidth]{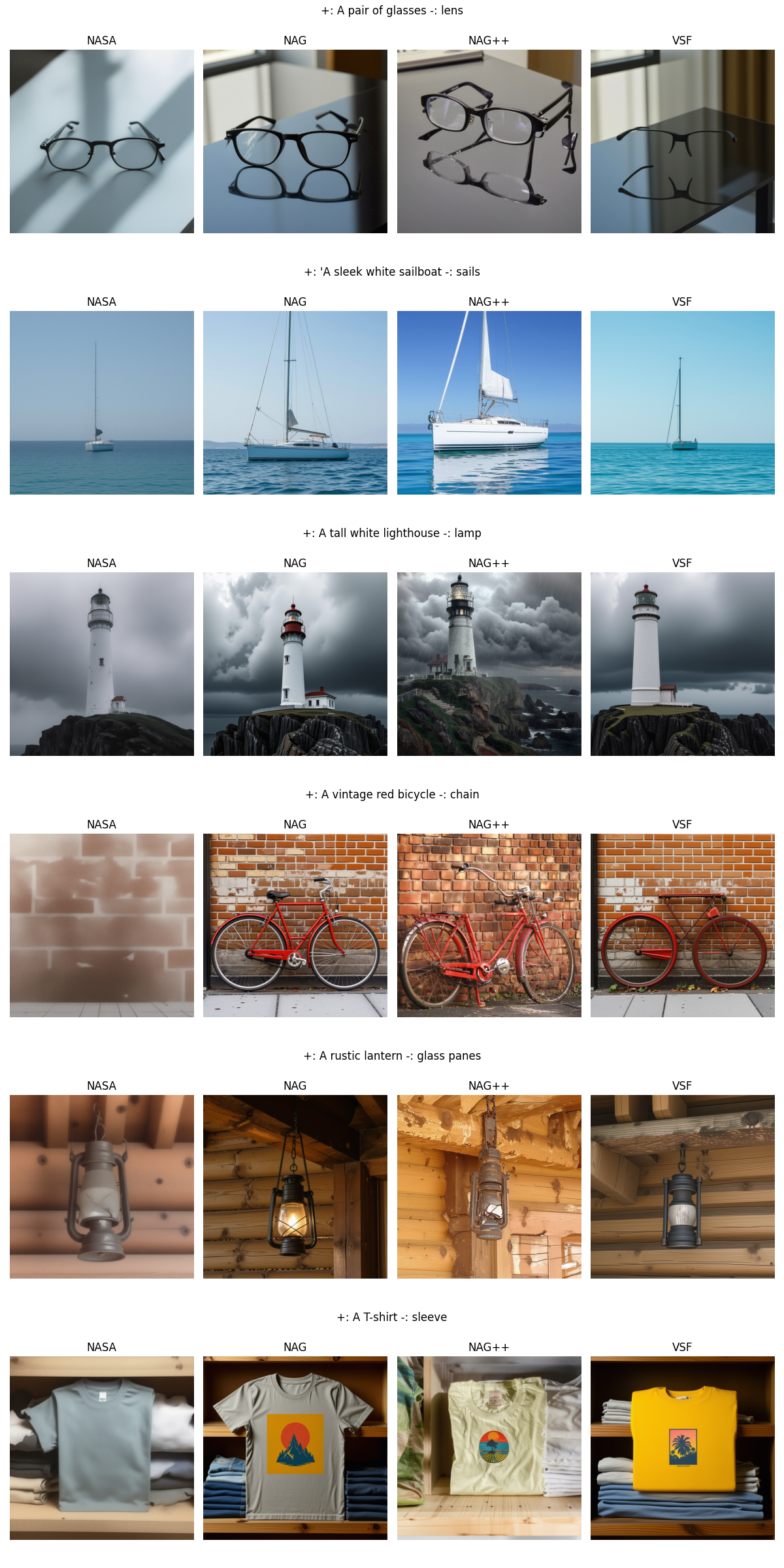}
    \caption{Selected Results for Comparison. Positive promtes are condensed for spacing.}
    \label{fig:res}
\end{figure}
Selected qualitative results are shown in Figure~\ref{fig:res}. The positive prompt is condensed for spacing. For the glasses without lens images, both NAG and NAG Strong generated classes clearly have a lens. For the VSF-generated image, we can see the lens is missing, even though the frames are floating. However, this issue was also presented in NAG Strong's image, even though it still has glasses. For a sailboat without sails, all other methods generated smaller but still existing sails, while VSF successfully avoided sails. In the third image of a lighthouse without a lamp, both VSF and NASA have no visible lamp, while the images from NAG and NAG Strong have a clear lit lamp. In the image of a bicycle without a chain, NASA generated a blurry image without bikes at all, while NAG generated a normal image, and NAG Strong generated a slightly distorted image of a bike with no seat yet the chain is still present. VSF successfully generated a bike without a chain, even though it also removed the seat. For the prompt of a lantern with no glass panes, NASA generated a lamp with frosted panes, NASA++ generated a classic glass pane, and NASA++ generated broken frosted panes. VSF, in this case, generated a pane that is clearly not glass. In the last example of a T-shirt, NASA generated a blurry image with still one sleeve visible, and NAG generated the T-shirt with both sleeves visible. NAG Strong and VSF both avoided the sleeve, even though NAG Strong has some artifacts.   

\section{Qualitative Results for Wan}
\begin{figure}
    \centering
    \includegraphics[width=0.9\linewidth]{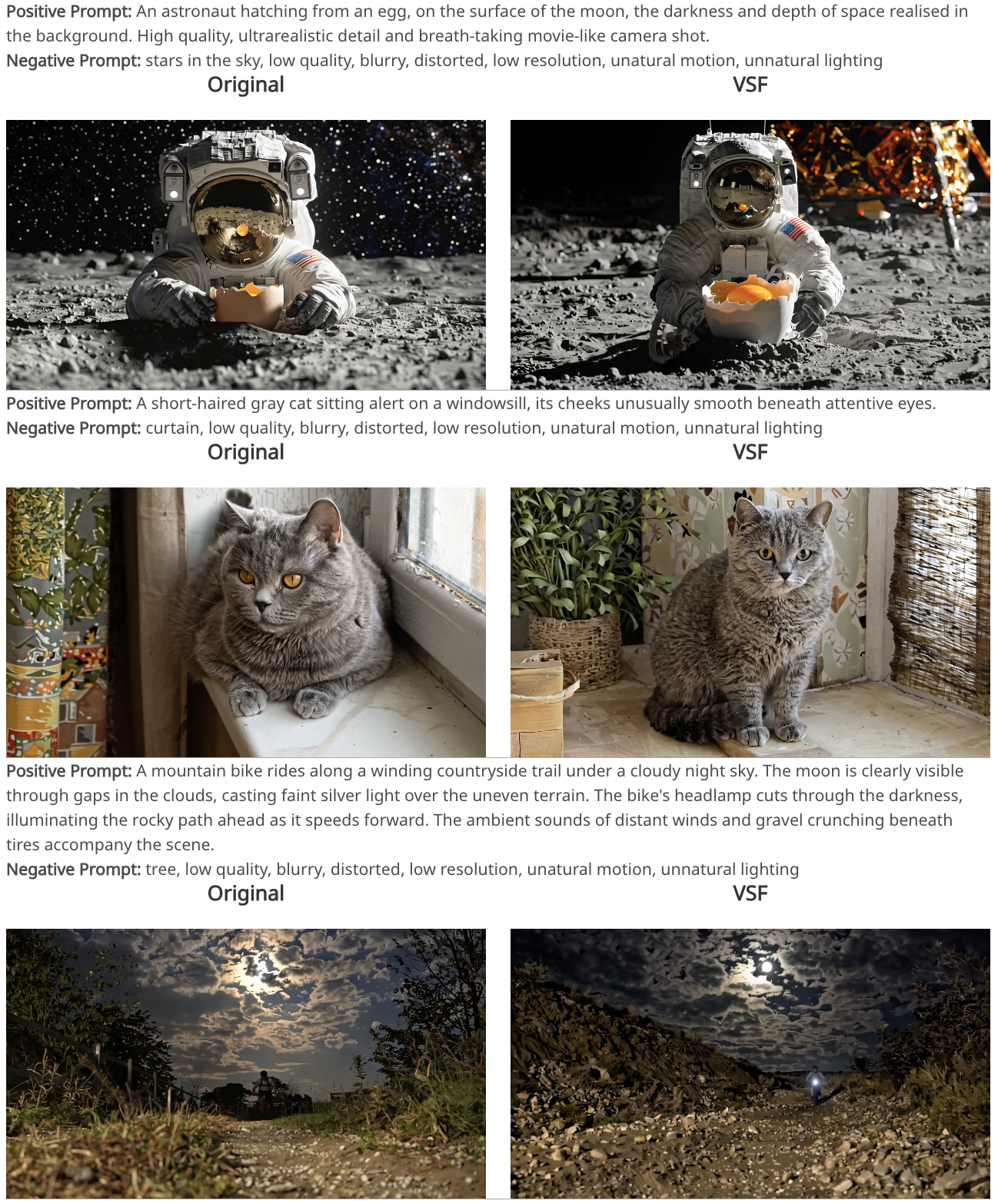}
    \caption{Qualitative Results for Wan}
    \label{fig:wan_res}
\end{figure}

In Figure~\ref{fig:wan_res}, we showed 3 examples generated from Wan-2.1-14B. In the first example, we successfully removed the stars in the background while keeping other elements intact. On the right side, in the absence of stars, the moon lander is generated to fill the space. In the second video, we successfully generated a windowsill without a curtain. In the last video, the generated video from VSF contains no trees on the left, and instead, it fills it with a hill. There are still some bushes on the right side, which do not violate the negative prompt of ``trees. All videos have the same high quality as the original one. 

\section{Failure Cases}
\begin{figure} 
    \centering
    \includegraphics[width=0.9\linewidth]{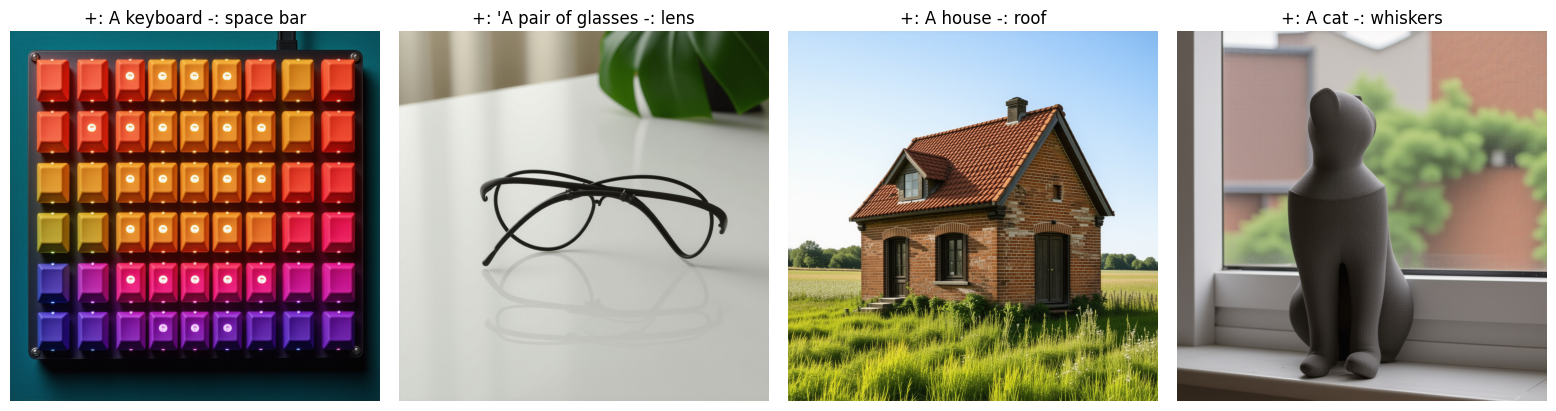}
    \caption{Failed examples, positive prompts are condensed for spacing.}
    \label{fig:failed}
\end{figure}
Like any method, our method is not perfect, especially in a challenge dataset like NegGenBench. In Figure~\ref{fig:failed}, we showed 4 failed cases. In the case where we want to generate a keyboard without a spacebar, the generated object is technically a key-board (an array of keys) and has no space bar, but it is not what people imagine when they think about "keyboard with no spacebar." The second failed case is another image from glasses with no lens; the generated image has no lens, but the frame is twisted in an unnatural way. In the third example, where a house with no roof is needed, VSF completely missed the negative prompt, possibly due to the strong association between roof and house. In the last example of a cat with no whiskers, the generated image technically has no whiskers and looks like a cat, but it looks more like a cat statue instead of a living cat. 
\section{Non-Object Negative Guidance}
In this paper, we focused on removing a critical component in the image. To further validate our negative guidance method in other areas, we also tested it on style avoidance. In Figure~\ref{fig:non_object} (in main text), we show four examples, each of which is generated using the same seed. We can see that when prompted with famous artwork (e.g.,"A painting of Starry Night from the 1890s" or "Mona Lisa oil painting") but with a negative prompt of the artist's name style, the generated image avoided any elements related to the style (including the town in the Starry Night) but kept the semantic meaning of the positive prompt. When prompted to give an old photo but not monochrome, the generated image is more like an old-style color photo, follows both non-monochrome and also not very bright (as old photos, even in color, are less vibrant). We find these examples interesting and think they can be used for machine unlearning, using a similar method as in \citep{gandikota_erasing_2023}.

\section{An Experiment on Anti-Aesthetics Arts}  
Current image generation models are typically finetuned to align with so-called ``human preference.'' However, we argue that there is no universal standard for human preference, and it cannot be defined solely by developers, who inevitably bring their own interests and assumptions. Aligning models exclusively with such values risks introducing bias and potentially marginalizing minority perspectives and interests \citep{arzberger_nothing_2024, TurchinManuscript-TURAAP, university_of_tartu_challenges_2020, guo_position:_2025}.   

In the context of image generation, this alignment may lead to homogenization of style or taste, producing only broadly pleasing outputs for the general population. Such uniformity can suppress niche demands for degraded, low-quality, or unconventional aesthetics. To counteract this, one possible approach is the use of negative guidance to steer outputs away from mainstream preferences. In this experiment, we tested how VSF can address this issue. We ran our VSF in settings where $\alpha=0$, which shows on the left, and $\alpha\in[0,4]$, which shows on the right side. The image with $\alpha=0$ might not be the same as the one without guidance, but should be an image without negative guidance.
\begin{figure}
    \centering
    \includegraphics[width=\linewidth]{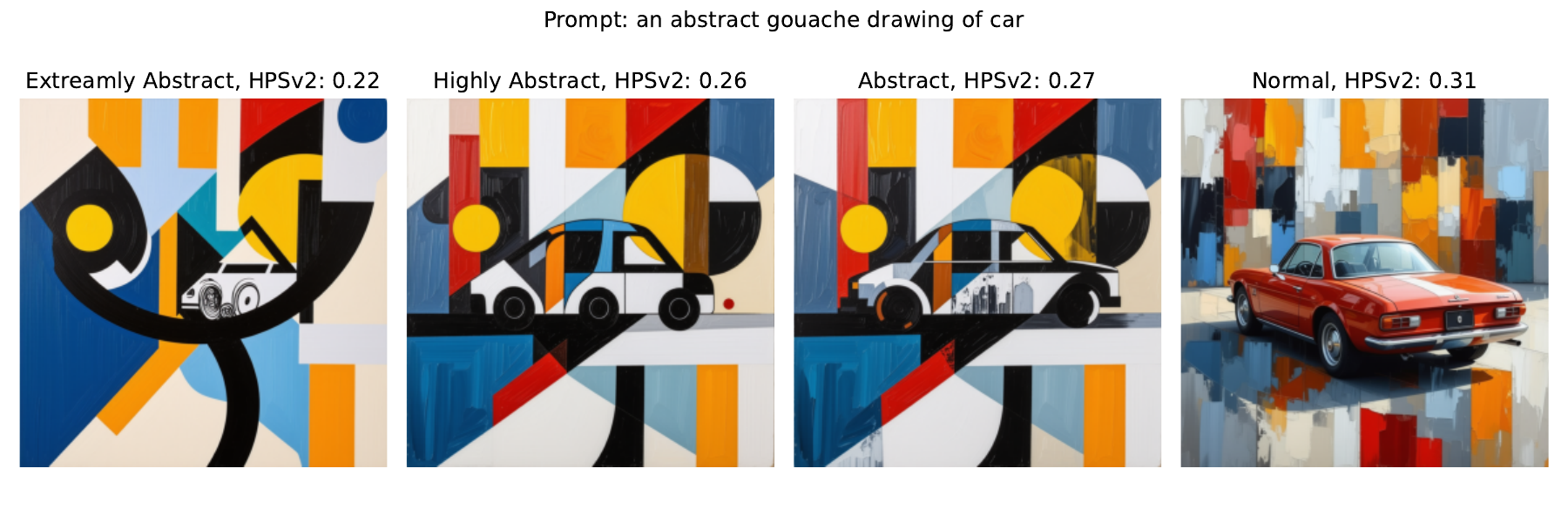}
        \caption{Image with abstract style receives a lower score in HPSv2 (reflecting human preference; traditionally, models aim for higher scores).}
    \label{fig:abstract_score}
\end{figure}
The first test used prompts containing the same object in both positive and negative form, with the goal of producing abstract art. This works by semi-canceling the main object, making it appear in an abstract form. Abstract styles are often disadvantaged in alignment settings, since reward models typically favor realistic or figurative outputs. VisionReward \citep{xu_visionreward_2025} encodes this bias through its scoring metric, and LAPIS \citep{maerten_lapis_2025} reports that abstract paintings generally receive lower preference scores. Figure~\ref{fig:abstract_score} shows that an abstract image gets a much lower score compared with a figurative one. As shown in the first two rows of Figure~\ref{fig:ugy_art}, the apple, people, and cat appear in abstract form, demonstrating a clear shift away from the default figurative tendency of the aligned models when VSF is applied. For the last image of a dog, we used ``cute'' as a negative prompt, which usually describes realistic objects, and we achieved a very abstract and artistic image. Figure~\ref{fig:abstract_scale} shows how abstract the image gets as the scale increases. More examples are shown in Figure~\ref{fig:abstract2}.

\begin{figure}
    \centering
    \includegraphics[width=\linewidth]{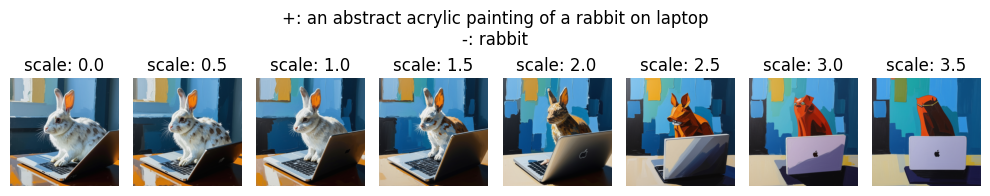}
    \caption{Abstraction of the image as scale increases.}
    \label{fig:abstract_scale}
\end{figure}

In the second test, the goal was to diverge from styles that are generally appreciated. The positive prompt specified the desired style, while the negative prompt contained descriptions of commonly preferred styles. Importantly, the positive prompt clearly described the intended output, so a faithful model should follow it rather than default to generalized human preference. The tested cases included desaturated color, sad emotion, pixelated art, insufficient lighting, unnatural colors, and a non-beautiful cat. Results show that the baseline model struggled to maintain these characteristics, often reverting to conventionally “beautiful” imagery, whereas VSF successfully produced outputs aligned with the specified unconventional styles.

\begin{figure}
    \centering
    \includegraphics[width=0.8\linewidth]{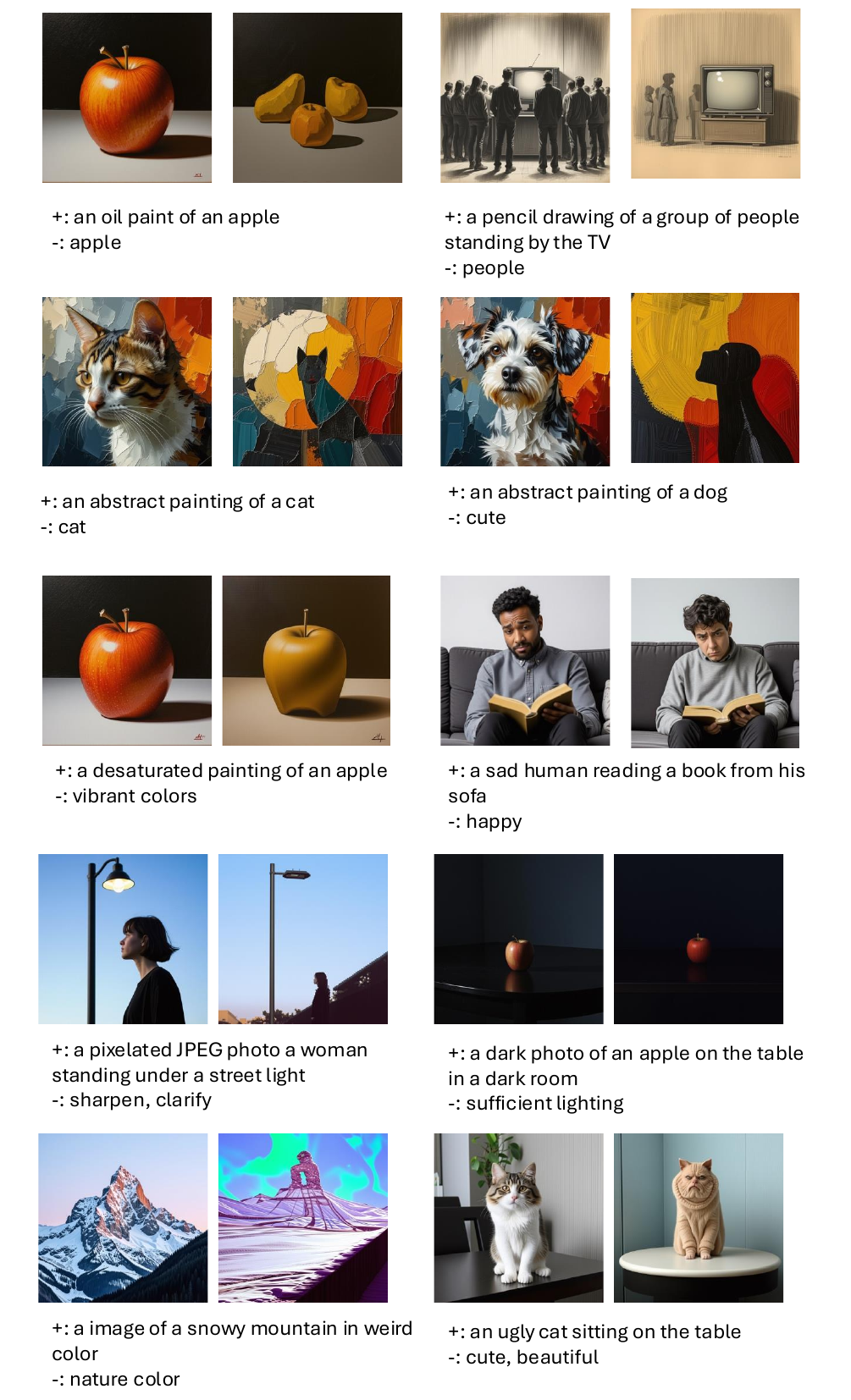}
    \caption{A test for anti-aesthetics. The left image is generated with $\alpha=0$, and the right image is generated with $\alpha>0$. These tests aim to move away from universally pleasing styles and demonstrate the ability to capture more diverse aesthetic preferences.}
    
    \label{fig:ugy_art}
\end{figure}

\section{Negation-Aware MLLM}
Upon visual inspection, we observed that GPT-4o effectively avoids many negative prompts, though occasionally the ambiguity within negative prompts (e.g., the term ``door" referring either to the door panel or the entrance) and the vision ability of MLLM itself (i.e., hallucination) leads to lower negative scores. It is possible that the negative scores for other methods might also be underestimated. We acknowledge this as a limitation associated with using an MLLM as the evaluator. Thus, we provided a better negotiation understanding of MLLM and used this MLLM to evaluate different guidance methods. 

To enable future research on evaluating negation prompting in generative models, a reliable and fast (LLaMA Maverick is too big) evaluation model is necessary. Previous CLIP-based studies concentrated on simple negations (e.g., "a cat that is not on the grass") rather than more complex cases such as those in NegGenBench. To address this, we finetuned a multimodal large language model, Qwen-2.5-VL-7B, called NegAwareQwen for improved understanding.

We created 100 additional prompts using GPT-5, each paired with a positive and a negative pair and 2 questions. For each prompt, we generated two images with the three models (NAG, VSF Quality, and NASA) and selected 722 for manual scoring on two dimensions: adherence to the negative prompt and overall quality. We did not assess the positive prompt evaluation as those are simpler, and almost all images in the dataset have a perfect positive prompt following. Negative adherence was rated on a three-level scale: 0 (ignored), 0.5 (partial), and 1 (fully followed). When generating the samples, we slightly randomly adjust the hyperparameter in a small range to create more diverse data (For VSF, $\alpha=3.3\pm1,\beta=0.2$; for NASA, $\alpha=0.15\pm0.05$; for NAG, $\phi=8\pm4,\alpha=0.5\pm0.2,\tau=4\pm2$). Since this makes image generation models generate sub-optimal images, the rating results of each model's images are not used for direct comparison. The dataset and the model will be opened after publication.

Note that here Llama showed a very weak r-score for quality; this is because all the images are evaluated using relatively high-quality images (unlike in the ablation study, where many images are lower quality). We did not compare the 7B untrained model because it often failed to output the structure data needed.  
\begin{table}[]
    \centering
    \small
    \caption{Negation-Aware LLM Evaluation on Testing Set}
    
        \begin{tabular}{l|llll}
        \hline
         & Parameters& r ($\uparrow$) & Acc ($\uparrow$) & F1 ($\uparrow$) \\\hline
        Llama Maverick & 400B & 0.05 & 0.83 & 0.59 \\
        Llama Maverick CoT & 400B & 0.03 & 0.77 & 0.50 \\
        Qwen-2.5-VL 32B & 32B & 0.28 & 0.80 & 0.36 \\
        Qwen-2.5-VL 32B CoT & 32B & 0.31 & 0.88& 0.70\\
        NegAwareQwen-7B& 7B & \textbf{0.37}& 0.86 & 0.65 \\
        NegAwareQwen-32B& 32B &  0.34&  \textbf{0.90}& \textbf{0.76}\\
        \hline
        \end{tabular}

    \label{tab:better_mllm}
\end{table}

The model was finetuned using prompts from the dataset of all 3 models. We trained the model using QLoRA \citep{dettmers_qlora:_2023} with rank of 8 and $r=8$, applied to query, key, value projections in both the vision encoder and language model with dropout of 0.1. Model is trained using $lr=5\times10^{-5}$ (with warm up and decay), WeightDecay=0.1, BatchSize=16, Epoch=5. The dataset is split into train-val with a 90-10 ratio based on the prompt level splitting and aiming for balanced scores in each split. We treat 0.5 as False and calculate negative scores as a binary metric. Results are presented in Table~\ref{tab:better_mllm} with comparison with the same model without finetuning and LLaMA-Maverick. 

\section{Evaluation Using NegAwareQwen}
We re-evaluated VSF Quality, NAG, NAG Strong, NASA, GPT-4o, and Nano Banana images generated using the prompts and seeds in the main text of the paper using our finetuned NegAwareQwen; the results are shown in Table~\ref{tab:res2}.  We did not round the 0.5 score. We used a positive score from the original Table~\ref{tab:res} as our finetuned version was trained on largely positive compliance samples. 
The results match our observation with human validation and MLLM evaluation, that our method gets the highest negative score while having better or comparable positive and quality scores with other open source guidance methods. 

\begin{table}[]
    \centering
        \caption{Evaluation of Different Methods using Our NegAwareQwen-32B} 
        \begin{tabular}{l|rrr}
        \hline
         & Positive Score ($\uparrow$) & Negative Score ($\uparrow$) & Quality Score ($\uparrow$)\\
        \hline
         
        VSF Quality & 0.980& 0.330 & \textbf{0.814}\\
        VSF Strong & 0.870& \textbf{0.415} & \textbf{0.814}\\
        NASA & 0.950& 0.224 & 0.727 \\
        NAG Strong & 0.950& 0.168 & 0.795 \\
        NAG & \textbf{1.000}& 0.147 & 0.812 \\
        \hline
        GPT-4o &  \textbf{0.978}&  \textbf{0.619}&  0.812\\
        Nano Banana & 0.985 &  0.406 & \textbf{0.817}  \\
        \hline
        
        \end{tabular}
        
    \label{tab:res2}
\end{table}
\section{Abliation Study On $\alpha$ and $\beta$}
\begin{figure}
    \centering
    \includegraphics[width=0.5\linewidth]{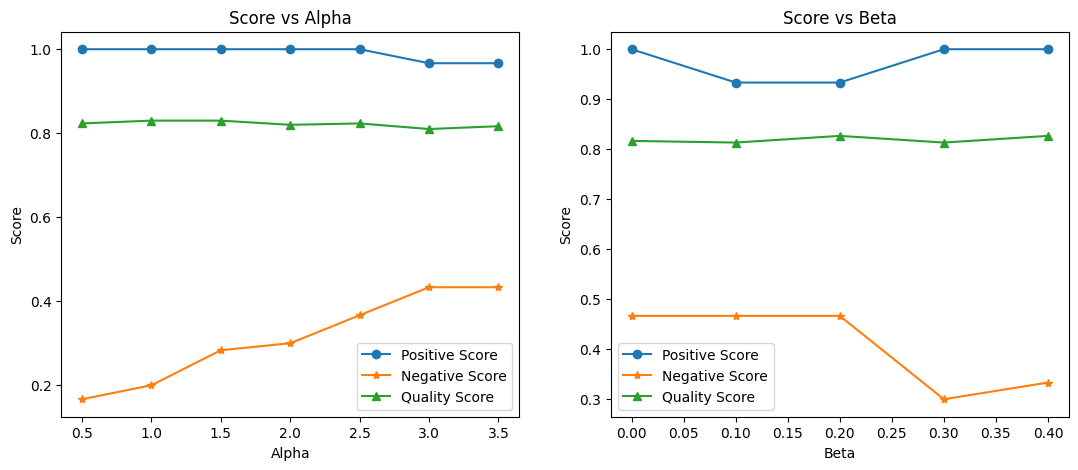}
    \caption{Relationships between metrics and $\alpha$ and $\beta$}
    \label{effects2}
\end{figure}

To study the effects of $\alpha$ and $\beta$ and hyperparameter sensitivity, we studied the effects of the two hyperparameters. We used 30 randomly selected prompts from the dataset and tested the effects of $\alpha$ and $\beta$ on the positive, negative, and quality scores. When testing the effects of $\alpha$, we set $\beta$ to 0, and when studying the effects of $\beta$, we set $\alpha$ to 3.5. All generations use the same seed. The images are evaluated using our NegAwareQwen-32B running for negative and quality score and used LLaMA for the positive score. Qualitative results of the effects of $\alpha$ and $\beta$ are shown in Figure~\ref{fig:effects} and quantitative results are shown in Figure~\ref{effects2}. We can see that as $\alpha$ increases, negative scores increase while positive scores decrease. When $\beta$ increases, the negative score decreases while the positive score increases, which could be noise. In both cases, the quality scores only change slightly.

\section{Future Work}
Future work may involve applying it to non-diffusion models (like Janus-4o \citep{chen_sharegpt-4o-image:_2025}) or models with complex text encoders (like Qwen-Image \citep{wu_qwen-image_2025}), improving robustness through normalization and blending techniques similar to those employed by NAG, and optimizing computational efficiency by using a better attention implementation. Additionally, we observed some inaccuracies in MLLM judgment due to ambiguities or minimal differences in visual differences. Conducting a larger-scale human evaluation study would help mitigate inaccuracies observed in MLLM-based assessments. Investigating the attention maps and diffusion trajectories of our model could further elucidate the underlying mechanisms of VSF. Decoupling the attention, such that it calculates the positive and negative attention separately and then uses the ratio to extrapolate the output, might yield better quality in exchange for runtime. Or, adding a scaling factor to the positive prompt for better control. 

\end{document}